\documentclass{article}

\usepackage{arxiv}

\usepackage[utf8]{inputenc} 
\usepackage[T1]{fontenc}    
\usepackage{hyperref}       
\usepackage{url}            
\usepackage{booktabs}       
\usepackage{amsfonts}       
\usepackage{nicefrac}       
\usepackage{microtype}      
\usepackage{lipsum}
\usepackage{graphicx}
\usepackage{longtable}
\usepackage{multirow}
\usepackage{rotating}
\usepackage{geometry}
\usepackage{afterpage}
\usepackage{xcolor}
\usepackage{pdflscape}
\usepackage{amsmath}

\usepackage[paper=portrait,pagesize]{typearea}

\title{\emph{NEMANet}: A CONVOLUTIONAL NEURAL NETWORK MODEL FOR IDENTIFICATION OF NEMATODES SOYBEAN CROP IN BRAZIL}

\author{Andre da Silva Abade \\
Federal Institute of Education \\
Science and Technology of Mato Grosso, Brazil \\
\texttt{andre.abade@bag.ifmt.edu.br} \\
\And
Lucas Faria Porto \\
Department of Mechanical Engineering \\
University of Brasilia - Brazil \\
\texttt{lucasfporto@gmail.com} \\
\And 
Paulo Afonso Ferreira \\
Department of Agronomy \\
Federal University of Mato Grosso, Brazil \\
\texttt{pafonsoferreira@yahoo.com.br}\\
\And 
Flavio de Barros Vidal \\
Department of Computer Science \\
University of Brasilia - Brazil \\
\texttt{fbvidal@unb.br} \\
}

\begin{document}
\maketitle

\begin{abstract}
Phytoparasitic nematodes (or phytonematodes) are causing severe damage to crops and generating large-scale economic losses worldwide. In soybean crops, annual losses are estimated at 10.6\% of world production. Besides, identifying these species through microscopic analysis by an expert with taxonomy knowledge is often laborious, time-consuming, and susceptible to failure. In this perspective, robust and automatic approaches are necessary for identifying phytonematodes capable of providing correct diagnoses for the classification of species and subsidizing the taking of all control and prevention measures. This work presents a new public data set called NemaDataset containing 3,063 microscopic images from five nematode species with the most significant damage relevance for the soybean crop. Additionally, we propose a new Convolutional Neural Network (CNN) model defined as {\it NemaNet} and a comparative assessment with thirteen popular models of CNNs, all of them representing the state of the art classification and recognition. The general average calculated for each model, on a from-scratch training, the NemaNet model reached 96.99\% accuracy, while the best evaluation fold reached 98.03\%. In training with transfer learning, the average accuracy reached 98.88\%. The best evaluation fold reached 99.34\% and achieve an overall accuracy improvement over 6.83\% and 4.1\%, for from-scratch and transfer learning training, respectively, when compared to other popular models.
\end{abstract}

\keywords{Plant Disease \and
Phytonematodes \and
Plant Pathogen \and
Plant-parasitic nematode \and
CNN \and
Deep Learning}

\section{Introduction} \label{sec:introduction}

The identification of plant diseases is one of the most basic and important agricultural activities. In cases, identification is performed manually, visually or by microscopy~\cite{ConceptsPlantPest:1995,Agrios:2005}. The main problem of the visual assessment to identify diseases is that the expert takes on a subjective task, prone to psychological and cognitive phenomena that can lead to prejudice, optical illusions, and, finally, error~\cite{BARBEDO:2016}. Furthermore, laboratory analysis, such as molecular-based approaches cultures, immunological or pathogens generally time-consuming, leaving provide answers in a timely manner~\cite{Kaur:2019}.

Plant-parasitic nematodes cause damage to crop plants on a global scale. Phytonematodes have caused significant losses to Brazilian agriculture, being considered among the primary pathogens in soybean crops. In addition to the losses, the control of phytonematodes is a challenge~\cite{Ferraz:2010,Freitas:2014,FerrazBrown:2016}.

These pathogens are considered to be hidden enemies of producers because it is not always possible to view or identify them in the field~\cite{Ferraz:2010,FerrazBrown:2016}. Symptoms in the aerial part of plants, in most cases, are easily confused with other causes, among them, nutrient deficiency, attack of pests and diseases, drought, and soil compaction~\cite{Coyne:2007}. According to the Brazilian Society of Nematology~\cite{SBN:2019}, losses vary on average between 5\% and 35\%, depending on the type of crops. In more severe cases, the losses can be even greater.

The basic premise of controlling these organisms is how species of nematodes are present in the crop's area. Currently, this identification is carried out by highly specialized technicians in laboratories, where the extraction of nematodes from soils and roots is performed~\cite{TihohoD:1997,Coyne:2007}. It should be noted that the vast majority of the methods used to identify and classify nematodes are made by visual display based on the morphological characters of the nematode and rely exclusively on the know-how of the specialist professional for proper recognition.

In this context, the development of automatic methods capable of identifying diseases quickly and reliably is convincing. Automated solutions for identifying plant diseases using images and machine learning, especially Convolutional Neural Networks (CNNs), have provided significant advances to maximize the accuracy of correct diagnosis~\cite{Kaur:2019,Thyagharajan:2019}.

Convolutional neural networks are a class of machine learning models currently state of the art in many computer vision tasks, including object classification and detection. Part of this success lies in the ability of a CNN to perform automated feature extraction, as opposed to classical methods that may require handcrafted features~\cite{LECUN:2015,ArchictetureCNN:2019}. In a sense, training a CNN to expert-level performance crystallizes some of a pathologist’s or crop scout’s diagnostic capabilities to be shared with any number of growers at any time or location.

However, it was impossible to find a public domain dataset with microscopic images that would allow classification of phytonematodes using CNNs, nor was it possible to locate studies that would allow a comparison of results and characterization of state of the art.

In this way, we can summarize the contributions of this paper, highlighting the following objectives:

\begin{enumerate}
    \item  Build a database of cataloged and labeled images, subject to applications of computer vision techniques and machine learning in the process of identification and automated classification of phytonematodes;
    
    \item  Analyze and evaluate the different techniques and methods of automated identification of phytonematodes using CNNs;

    \item  Provide a benchmark for the most popular CNNs in the literature, showing the inference capacity of each model in the process of identifying the species of nematodes investigated.

\item Present our CNN architecture, called NemaNet, which overcame the results of traditional CNN architectures in the phytonematodes classification process.
    
\end{enumerate}

The organization of this paper is as follows: Section~\ref{sec:review} discusses the literature review of this paper. Next, Section~\ref{sec:RelWorks} discusses the related work of this paper. In Section~\ref{sec:methodology}, we describe the material and methods used in our approach, where we detail the construction of the dataset called NemaDataset, define the process of evaluation and analysis of CNNs presented by the literature, and propose a new architecture called NemaNet that innovates state of the art in the classification process of phytonematodes. In Section~\ref{sec:results}, the experimental results are presented. Next, in Section~\ref{sec:discussion}, the results are further discussed. Finally, Section~\ref{sec:conclusion} concludes this paper.

\section{Literature Review} \label{sec:review}

Soybean (\textit{Glycine max}) is a plant belonging to the legume family considered one of the oldest agricultural products known to humankind. Some reports reveal that soy planting dates back to 2,838 B.C. in China. Brazil ranks among the world's largest producers, both in planted area and in productivity in kilograms per hectare~\cite{USDA:2020,CONAB:2020}.

Ubiquitous in nature, phytoparasitic nematodes are associated with nearly every important agricultural crop and represent a significant constraint on global food security. In soybean crops, the main phytonematodes are root-knot nematodes (\textit{Meloidogyne} spp.), soybean cyst nematodes (\textit{Heterodera glycines}), and lesion nematodes (\textit{Pratylenchus} spp.) rank at the top of the list of the most economically and scientifically important specimens due to their intricate relationship with the host plants, wide host range, and the level of damage ensued by infection. However, there are other nematodes associated with soybean crops such as \textit{Helicotylenchus dihystera}, \textit{Mesocriconema} ssp., \textit{Rotylenchulus reniformis}, \textit{Scutellonema brachyurus}, \textit{Trichodorus} spp.,  \textit{Aphelenchoides besseyi}, \textit{Tubixaba} spp. among others~\cite{TihohoD:1997,FerrazBrown:2016,Bernard:2017}.

Nematodes are extremely abundant and diverse animals; only insects exceed their diversity. Most nematodes are free-living and feed on bacteria, fungi, protozoans, and other nematodes (40\% of the described species); many are parasites of animals invertebrates and vertebrates (45\% of the described species) and plants (15\% of the described species)~\cite{Lambert:2002}.

Plant-parasitic nematodes occur in all sizes and shapes. The typical nematode shape is a long and slender worm-like animal, but the adult animals are often swollen and no longer resemble worms. The phytonematodes are small, 300 to 1,000 micrometers,  with some up to 4 millimeters long, by 15–35 micrometers wide~\cite{Lambert:2002,Agrios:2005}.

Identification of nematodes to the species level often requires detailed morphological analysis, growth of the nematode on different host plants, or DNA or isozyme analysis~\cite{Dropkin:1989,Lambert:2002}. Common morphological features used in nematode identification include the mouth cavity (presence or absence and shape of a stylet), the shape and overlap of the pharyngeal glands with the intestine, size and shape of the nematode body at the adult stage, size of the head, tail, and number and position of ovaries in the female. More subtle characters may include the number of lines on the nematode’s cuticle or the presence or absence of pore-like sensory organs~\cite{Lambert:2002,Perry:2011}.

Nematodes are considered one of the most difficult pathogens to be identified, either due to their small size or difficulty observing key characteristics for diagnosis under conventional light microscopy. The morphological and morphometric differences are relatively small. They require considerable knowledge in taxonomy for a safe determination of the species, corroborating it as a complicating factor for the detection and correct identification of these agents in routine analyzes~\cite{Tarique:2015,Bernard:2017}. Thus, it is clear that methods of detection and identification of pathogens, based on morphological aspects implemented exclusively by visual analysis by specialists, often do not meet the needs of quality control systems at the level of laboratory routine.

Furthermore, unfortunately, there is a very shortage of nematologists with taxonomy training, primarily due to the decrease in the number of qualified professionals available on the market. Added to these difficulties is the fact that meeting the demand for a large volume of analyzes, in short periods, causes the use of some health tests to generate a high number of false positives in the process of identification and population counting of nematodes present in the sample.

\subsection{Microscopic Image Analysis}

Microscopes have long been used to capture, observe, measure, and analyze various living organisms' images and structures at scales far below average human visual perception. With the advent of affordable, high-performance computer and image sensor technologies, digital imaging has come into prominence and is replacing traditional film-based photomicrography as the most widely used microscope image acquisition and storage method. Digital image processing is not only a natural extension but is proving to be essential to the success of subsequent data analysis and interpretation of the new generation of microscope images~\cite{Microscope:2008}.

Limitations of optical imaging instruments and the noise inherent in optical imaging make the image enhancement process desirable for many microscopic image processing applications. Image enhancement is the process of enhancing the appearance of an image or a subset of the image for better contrast or visualization of certain features and subsequently facilitating more accurate image analysis. With image enhancement, the visibility of selected features in an image can be improved, but the inherent information content cannot be increased. Thus, the challenge lies not in processing images but in processing them correctly and effectively~\cite{Microscope:2008}. 

More often than not, the images produced by a microscope are converted into digital form for storage, analysis, or processing before display and interpretation~\cite{MicGuide:2016}. Digital image processing significantly enhances the process of extracting information about the specimen from a microscope image. For that reason, digital imaging is steadily becoming an integral part of microscopy. Digital processing can be used to extract quantitative information about the specimen from a microscope image, and it can transform an image so that a displayed version is much more informative than it would otherwise be~\cite{MicComp:2019}.

In this perspective, Morphological image Processing (MP) is based on probing an image with a structuring element and either filtering or quantifying the image according to how the structuring element fits (or does not fit) within the image. A binary image is made up of foreground and background pixels, and connected sets of foreground pixels make up the objects in the image. Morphological processing has applications in such diverse areas of image processing as filtering, segmentation, and pattern recognition, to both binary and grayscale images. One of the advantages of MP is being well suited for discrete image processing because its operators can be implemented in digital computers with complete fidelity to their mathematical definitions. Another advantage of MP is its inherent building block structure, where complex operators can be created by the composition of a few primitive operators~\cite{Microscope:2008}. 

Nevertheless, microscopists and biologists are flooded with data that they have to normalize, filter, denoise, deblur, reconstruct, register, segment, classify, etc. One of the problems inherent in light microscopy is the generation of noise in the images. Noise is random fluctuations in the intensity in image pixels that obscure the real signal generated along the sample step. Noise is always present in images due to low light conditions, collecting limited numbers of photons, and the electric circuitry of the microscope~\cite{MicComp:2019}.

Microscope imaging and image processing are of increasing interest to the scientific and engineering communities. Recent developments in cellular, molecular and nanometer-level technologies have led to rapid discoveries. They have significantly advanced the frontiers of human knowledge in biology, medicine, chemistry, pharmacology, and many related fields.

\subsection{Convolutional Neural Networks}\label{subsec-cnn}

Computer Vision, along with Artificial Intelligence (AI), has been developing techniques and methods for recognizing and classifying objects with significant advances~\cite{ARNAL:2013}. According to ~LeCun et al. (2015)\cite{LECUN:2015} and, deep learning allows computational models to learn representations of data with multiple levels of abstraction, improving the state-of-the-art in many domains, such as speech recognition, object recognition, object detection.

The simplest Deep Learning models are called Deep Feedforward, in which information is only propagated in one direction through neurons. Other examples of algorithms are: Back-Propagation, Convolutional Neural Network (CNN), Recurrent Neural Network (RNN), including Long Short-Term Memory (LSTM) and Gated Recurrent Units (GRU), Auto-Encoder (AE), Deep Belief Network (DBN), Generative Adversarial Network (GAN), and Deep Reinforcement Learning (DRL)~\cite{Goodfellow:2016}. One particular type of deep, feedforward network that was much easier to train and generalized much better than networks with full connectivity was the convolutional neural networks~\cite{Vidal:2019}.

\begin{figure}[!htpb]
  \centering
  \includegraphics[width=1\columnwidth]{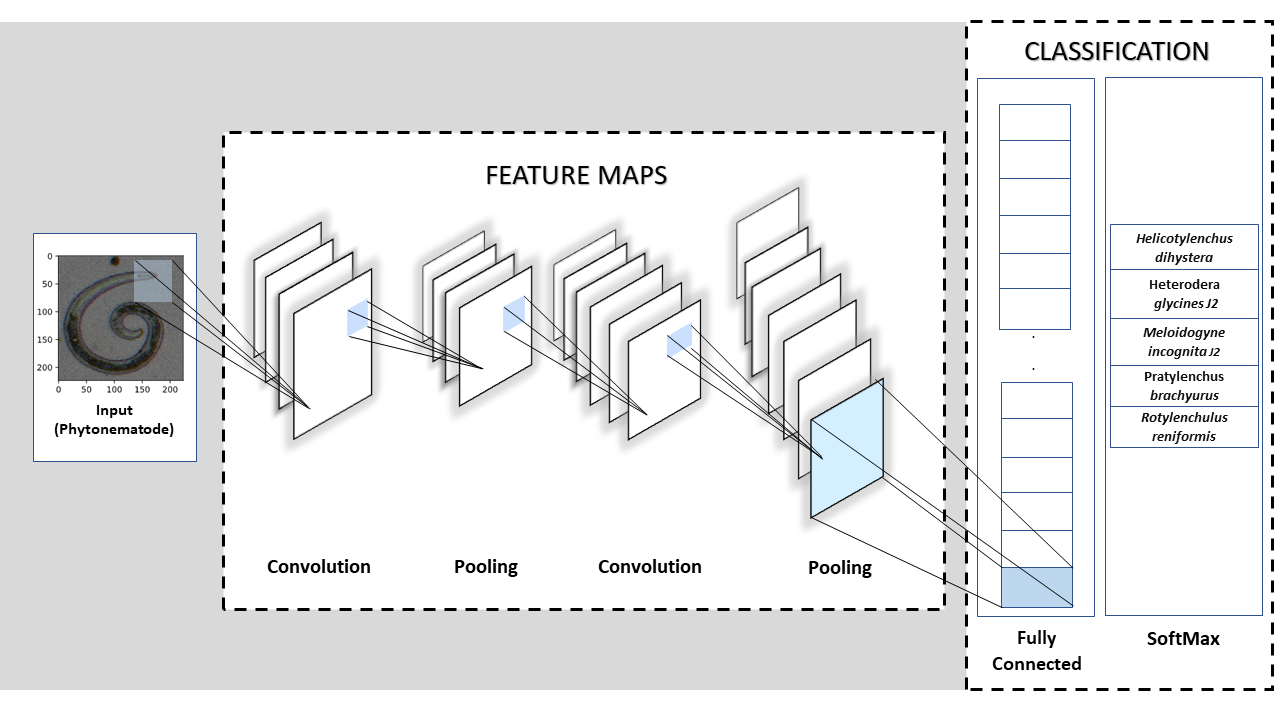}
  \vspace{-.65cm}
\caption{A Conceptual Architecture of a Convolutional Neural Network (CNN)}\label{fig:CNN}
\end{figure}

The CNNs constitute one of the most powerful techniques for modeling complex processes and performing pattern recognition in applications with a large amount of data and pattern recognition in images (\cite{LECUN:2015}). This one is a connectionist approach that stands out as one of the most prominent because it allows the automatic extraction of features. Their results in some experiments are already superior to humans in large-scale reconnaissance tasks. Figure~\ref{fig:CNN} presents a conceptual architecture for convolutional neural networks.

In recently researches of Convolutional Neural Networks many approaches already make use of popular architectures such as LeNet~\cite{LeNet:1998}, AlexNet~\cite{AlexNet:2012}, VGGNet~\cite{VGGNet:2014}, GoogLeNet~\cite{GoogLeNet:2015}, InceptionV3~\cite{InceptionV3:2016}, ResNet~\cite{ResNet:2016} and DenseNet~\cite{DenseNet:2017}, considerably increasing the accuracy in the identification and recognition of objects. New topologies and architectures dawn for many others approaches, as described in~\cite{AndaVidal-LCNN}.

\section{Related Works} \label{sec:RelWorks}

When conducting a literature review on related works on identifying phytonematodes by species using microscopic images, although few approaches use convolutional neural networks in the classification process, several studies have already addressed the need and proposed some solutions.

Silva et al.(2003)\cite{identNema01:2003} developed a hybrid system combining image processing, mathematical morphology, and artificial neural network techniques to detect the characteristics of phytonematodes. However, his study does not clarify how to correct the problems caused by lighting noise characteristic of microscopic images.

A study proposed by Doshi et al. (2007)\cite{identNema04:2007} uses hyperspectral data to identify two species, namely \textit{Meloidogyne incognita} and \textit{Rotylenchulus reniformis} applying methods based on Discrete Wavelet Transform (DWT) and Self-Organized Maps (SOM). The authors explore the possibility of combining these two methods of feature extraction and dimensionality reduction to improve the classification of these pathogens. The results demonstrate an accuracy for individual and general classification between the two nematode species, with a 95\% confidence interval, using a supervised SOM classification method.

Rizvandi et al. (2008)\cite{identNema02:2008} developed an algorithm capable of detecting the length of nematodes by separating them into small blocks and calculating the angle direction of each block. This approach solves the problem of nematodes superimposed with other nematodes. The proposed method is a vision algorithm with a 7.9\% of False Rejection Rate (FRR) and 8.4\% False Acceptance Rate (FAR) on a database of $255$ isolated and overlapped worms. 

The authors in \cite{identNema03:2013} presented an application called ``WormSizer'' that calculates the dimensional characteristics of nematodes, such as size, trajectory, etc. Nevertheless, the authors do not detail, in the results, the different situations in which nematodes can be found, such as the example of nematodes overlapping each other and the variety of species existing in the same sample.

A CNN-based regression network that can produce accurate edge detection results is proposed by Chou et al. (2017)\cite{identNema06:2017}. The feature-based mapping rules of the network are learned directly from the training images and their accompanying ground truth. Experimental results show that this architecture achieves accurate edge detection and is faster than other CNN-based methods.

The proposed system by Toribio et al. (2018)~\cite{identNema07:2018} is an algorithm oriented to detect the physical characteristics of nematodes in tropical fruit crops (width and length). The algorithm involves image acquisition of nematodes through a microscope, obtain the luminance component of the image, illumination correction, binarization by histogram, object segmentation, discrimination by area, surface
detection, and calculation of the Euclidean distance to approximate the physical characteristics of specimens in a sample of nematodes. Favorable results were obtained in detecting the features of the juvenile of the second stage (J2) of a species of \textit{Meloidogyne}. The results were validated from those obtained by clinical analysis of the specialists, achieving up to 85\% of success. 

An approach proposed by Liu et al. (2018)\cite{identNema11:2018} presents to use a deep convolutional neural network image fusion based multilinear approach for the taxonomy of multi-focal image stacks. A deep CNN-based image fusion technique is used to combine relevant information of multi-focal images within a given image stack into a single image, which is more informative and complete than any single image in the given stack. The experimental results on nematode multi-focal image stacks demonstrated that the deep CNN image fusion-based multilinear classifier reaches a higher classification rate (95.7\%) than the previous multilinear-based approach (88.7\%).

The authors in Chen et al. (2019)~\cite{identNema08:2019} proposed an algorithm, LMBI (Local Maximum of Boundary Intensity), to propose instance segmentation candidates. In a second step, an SVM classifier separates the nematode cysts among the candidates from soil particles. On a dataset of soil sample images, the LMBI detector achieves near-optimal recall with a limited number of candidate segmentations. The combined detector/classifier achieves recall and precision of 0.7.

An approach by Aragón et al. (2019)~\cite{identNema09:2019} proposes an algorithm oriented towards detecting the amount of damage or infection caused by the\textit{Meloidogyne incognita} nematode through the extraction of physical features in digital images of vegetable roots.  The algorithm consists of a thresholding step, a filtering stage, labeling, and physical feature extraction. Next, the obtained data feeds a neural network, which determines the infection level through the Zeck-scale. Results showed a 98.62\% specificity level and a 93.75\% sensitivity level.

Chen et al. (2020)~\cite{identNema10:2020} propose a framework for detecting worm-shaped objects in microscopic images based on convolutional neural networks. The trained model predicts worm skeletons and body endpoints. With light-weight backbone networks, we achieve 75.85\% precision, 73.02\% recall on a potato cyst nematode dataset, and 84.20\% precision, 85.63\% recall on a public \textit{Caenorhabditis elegans} dataset.

\section{Material and Methods} \label{sec:methodology}

\begin{figure}[!htpb]
  \centering
  \includegraphics[width=0.65\columnwidth]{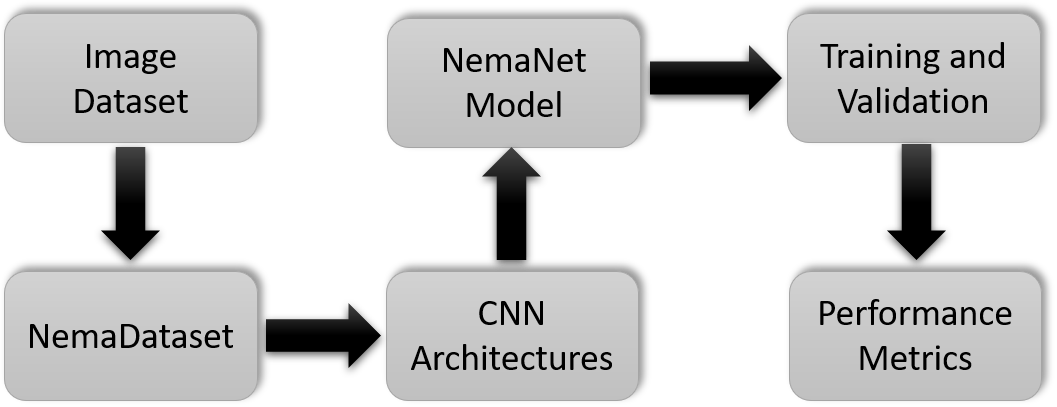}
\caption{The workflow of the method for the identification of nematodes from the soybean crop using NEMANet Convolutional Neural Network.}\label{fig:workflow}
\end{figure}

The workflow for the proposed method is described in Figure~\ref{fig:workflow}, and the following subsections present each detail involved in the process of building our approach. 

\subsection{Image Dataset}

We created a dataset called NemaDataset containing 3,063 microscopic images of the five species of phytonematodes with greater damage relevance to soybean crops\footnote{More information to Download is available in (double blind revision info.)}. The process of extracting the pathogens in the samples of plants and soil was made following the protocols proposed by Coolen and D'herde (1972)~\cite{Coolen:1972}, and Jenkins (1964)\cite{Jenkins:1964}. After the extraction, with a light microscope, with 10$\times$ ocular lenses, 5$\times$ objective lenses, digital camera for a microscope with Panasonic sensor of 16 megapixels, 2.33 inches, CMOS and 21$\times$ magnification (ocular) coupled. Phytonematological analyses were carried out to identify genera and species based on morphological and morphometric characteristics of females and males organs and body regions. All images were separated and classified according to each species in a specific class (e.g., directory structure), where each captured image with a spatial dimension of 5120$\times$3840 pixels at 72 dpi. It is noteworthy that the settings for capturing images are identical to the routine laboratory protocol for identification and classification by nematological analysis.  In this way, the set of images captured replicates the same visual conditions obtained by a specialist in a routine laboratory. Table~\ref{Tab:dataset} shows the distribution of images by species.

\begin{table}[!htpb]
\caption{Image dataset composition with the species of  phytonematodes}
\label{Tab:dataset}
	\fontencoding{T1}
    \fontfamily{\sfdefault}
    \fontseries{m}
    \fontshape{n}
    \fontsize{9}{12}
    \selectfont
    \setlength{\tabcolsep}{7pt}
\centering
\begin{tabular}{lccc}
\hline
\multicolumn{1}{c}{\begin{tabular}[c]{@{}c@{}}Class\\ (species)\end{tabular}} & \multicolumn{1}{l}{Nº of Images} & \multicolumn{1}{l}{\begin{tabular}[c]{@{}l@{}}Resolution\\ pixel / dpi\end{tabular}} & \multicolumn{1}{l}{Magnification} \\ \hline
\textit{Helicotylenchus dihystera}                                                           & 556                              & \multirow{5}{*}{\begin{tabular}[c]{@{}c@{}}5120$\times$3840\\ 72 dpi\end{tabular}}        & \multirow{5}{*}{105$\times$}             \\
\textit{Heterodera glycines} (J2)                                                      & 605                              &                                                                                      &                                   \\
\textit{Meloydogine incognita} (J2)                                                    & 635                              &                                                                                      &                                   \\
\textit{Pratylenchus brachyurus}                                                       & 635                              &                                                                                      &                                   \\
\textit{Rotylenchulus reniformis}                                                      & 632                              &                                                                                      &                                   \\ \hline   \hline                                                                                     
\end{tabular}
\end{table}

\begin{figure}[!ht]
  \centering
  \includegraphics[width=0.85\columnwidth]{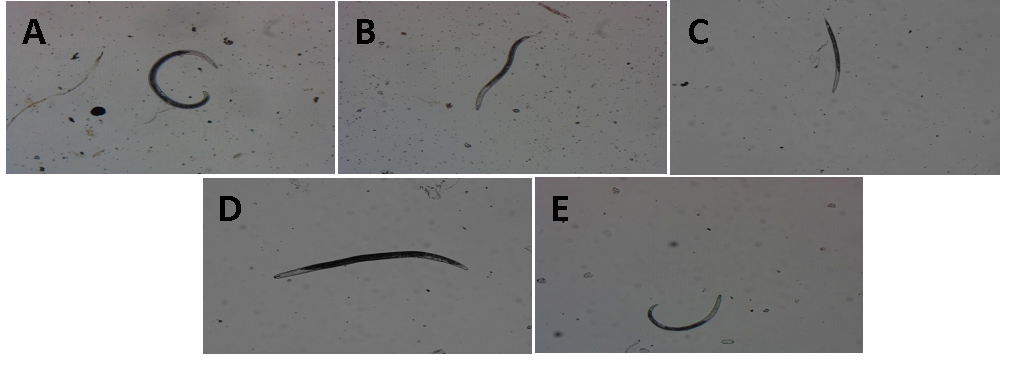}
  \vspace{-.65cm}
\caption{Examples raw images of phytonematodes NemaDataset: \textbf{(A)} \textit{Helicotylenchus dihystera} \textbf{(B)} \textit{Heterodera glycines} (J2). \textbf{(C)} \textit{Meloydogine incognita} (J2). \textbf{(D)} \textit{Pratylenchus brachyurus}. \textbf{(E)} \textit{Rotylenchulus reniformis}.  }\label{fig:samplesNemas}
\end{figure}

Because of the parasitic characteristics of the investigated phytonematodes, for the species \textit{Helicotylenchus dihystera} (ectoparasite) and \textit{Pratylenchus brachyurus} (endoparasite), all images were captured indistinctly. All of them maintain a wormlike pattern throughout their life cycle. As for the reniform nematode of the species \textit{Rotylenchulus reniformis}, the images were captured only of juvenile and sexually immature female pathogens. The sedentary Rotylenchulus females (semiendoparasite) were not interested in capturing and classifying the images, considering the complexity of handling for sample extraction. For the phytonematodes \textit{Heterodera glycines} and \textit{Meloidogyne incognita}, the images were captured only of juvenile pathogens of the second stage (J2). It was considering that this is its infective form, and all control measures should be done directed~\cite{NemaSojaEmbrapa:2010}. In Figure \ref{fig:samplesNemas}, we present five images captured by digital microscopy, representing the species investigated.


\vspace{-.45cm}

\subsection{NemaDataset pre-processing}\label{subsec-Nemdataset}

The optical components of the microscope act to create an optical image of the phytopathogen on the image sensor, which, these days, is most commonly a charge-coupled device (CCD) array. The optical image is a continuous distribution of light intensity across a two-dimensional surface. This two-dimensional projection is limited in resolution and is subject to distortion and noise introduced by the imaging process. The primary factors that can degrade an image in the digitizing process are, as follows: loss of detail,  noise,  aliasing, shading, photometric nonlinearity, and geometric distortion~\cite{Microscope:2008,MicComp:2019}. If each of these is kept low enough, then the digital images obtained from the microscope can be used for the training process of CNNs networks, guaranteeing quality in the level of inference.

Basically, in this pre-processing step, our efforts were focused on preserving a suitably high level of detail and signal-to-noise ratio while avoiding aliasing and doing so with acceptably low levels of shading, photometric nonlinearity, and geometric distortion. Additionally, we defined a region of interest in the images, making the pathogen centralized in the input images for training the CNNs network.

\begin{figure}[!ht]
  \centering
  \includegraphics[width=0.85\columnwidth]{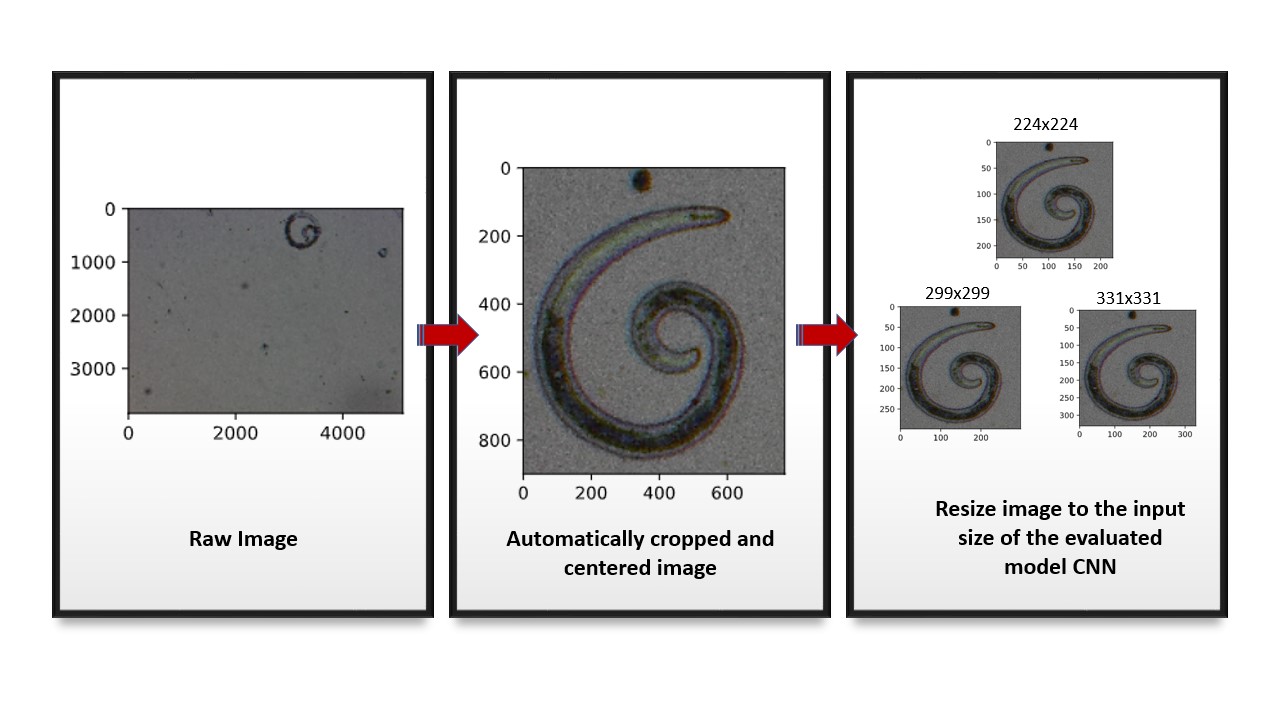}
  \vspace{-.95cm}
  \caption{Overview of pre-processing steps - Nemadataset}\label{fig:preprocess}    
\end{figure}

In Figure~\ref{fig:preprocess}, we demonstrate the evolution of the pre-processing steps of the raw images captured by the camera adapted to the microscope. From the raw image, a process of cropping and centralizing the object of interest was applied to significantly reduce features that are not determinants of the pathogen's classification.

\subsection{CNN Architectures}

Convolution network has been applied with great success for high-level computer vision tasks such as object classification and recognition. Recent studies demonstrated that they could also be used as a general method for low-level image processing problems, such as denoising and restoration. 

There are many CNN architectures proposed in the literature. In this study, we evaluated the performance of the 13(thirteen) popular model architectures provided by the Keras\footnote{More info.: https://keras.io/.} API. In Table~\ref{Tab:CNNModels}, we presented each model with its main characteristics.

\begin{table}[!ht]
\caption{Popular architectures evaluated }
\label{Tab:CNNModels}
	\fontencoding{T1}
    \fontfamily{\sfdefault}
    \fontseries{m}
    \fontshape{n}
    \fontsize{7}{9}
    \selectfont
    \setlength{\tabcolsep}{4pt}
\centering
\begin{tabular}{lrrclccccc}
\hline
\multicolumn{1}{c}{\multirow{2}{*}{Model}} &
  \multicolumn{1}{c}{\multirow{2}{*}{Size}} &
  \multicolumn{1}{c}{\multirow{2}{*}{Parameters}} &
  \multirow{2}{*}{Depth} &
  \multicolumn{1}{c}{\multirow{2}{*}{\begin{tabular}[c]{@{}c@{}}Image\\ Size\\ (H$\times$W)\end{tabular}}} &
  \multicolumn{5}{c}{Hyper parameters} \\ \cline{6-10} 
\multicolumn{1}{c}{} &
  \multicolumn{1}{c}{} &
  \multicolumn{1}{c}{} &
   &
  \multicolumn{1}{c}{} &
  \begin{tabular}[c]{@{}c@{}}Optimization \\ algorithm\end{tabular} &
  \begin{tabular}[c]{@{}c@{}}Batch\\ size\end{tabular} &
  Momentum &
  \begin{tabular}[c]{@{}c@{}}Weight\\ decay\end{tabular} &
  \begin{tabular}[c]{@{}c@{}}Learning\\ Rate\end{tabular} \\ \hline \hline
Xception &
  88 MB &
  22,910,480 &
  126 &
  229$\times$229 &
  \multirow{13}{*}{SGD} &
  32 &
  \multirow{13}{*}{0.9} &
  \multirow{13}{*}{\begin{tabular}[c]{@{}c@{}}1e-5\\ $\sim$ \\1e-6\end{tabular}} &
  \multirow{13}{*}{\begin{tabular}[c]{@{}c@{}}Base lr = 0.001\\ \\ Max lr = 0.00006\\ \\ Step size = 100\\ \\ Mode = triangular\end{tabular}} \\
VGG16             & 526 MB & 138,357,544 & 23  & 224$\times$224 &  & 32  &  &  &  \\
InceptionV3       & 92 MB  & 23,851,784  & 159 & 299$\times$299 &  & 32  &  &  &  \\
ResNet50          & 98 MB  & 25,636,712  & -   & 224$\times$224 &  & 100 &  &  &  \\
ResNet101         & 171 MB & 44,707,176  & -   & 224$\times$224 &  & 64  &  &  &  \\
ResNet152         & 232 MB & 60,419,944  & -   & 224$\times$224 &  & 64  &  &  &  \\
InceptionResNetV2 & 215 MB & 55,873,736  & 572 & 299$\times$299 &  & 32  &  &  &  \\
DenseNet121       & 33 MB  & 8,062,504   & 121 & 224$\times$224 &  & 32  &  &  &  \\
DenseNet169       & 57 MB  & 14,307,880  & 169 & 224$\times$224 &  & 32  &  &  &  \\
DenseNet201       & 80 MB  & 20,242,984  & 201 & 224$\times$224 &  & 32  &  &  &  \\
EfficientNetB0    & 29 MB  & 5,330,571   & -   & 224$\times$224 &  & 100 &  &  &  \\
EfficientNetB3    & 48 MB  & 12,320,535  & -   & 320$\times$320 &  & 32  &  &  &  \\
NASNetLarge       & 343 MB & 88,949,818  & -   & 331$\times$331 &  & 16  &  &  &  \\ \hline \hline
\end{tabular}
\end{table}

\subsection{The NemaNet Model}

Feature extraction and classification were performed on the microscope images extracted in the Data preprocessing step(~\ref{subsec-Nemdataset}) with a new model called NemaNet. The NemaNet model is based on a CNN topology. It includes an original DenseNet~\cite{DenseNet:2017} structure with multiple Inception blocks combining and optimizing two well-known architectures: InceptionV3 and DenseNet121 for deep feature extraction. The number of Inception blocks is used to control the depth, size, and the number of the model parameters. In this way, these blocks allow using multiples types of the filter size, instead of being restricted to a single filter size, in a single image block, which we concatenate and pass onto the next layer.

\begin{figure}[!htpb]
  \centering
  \includegraphics[width=0.42\columnwidth]{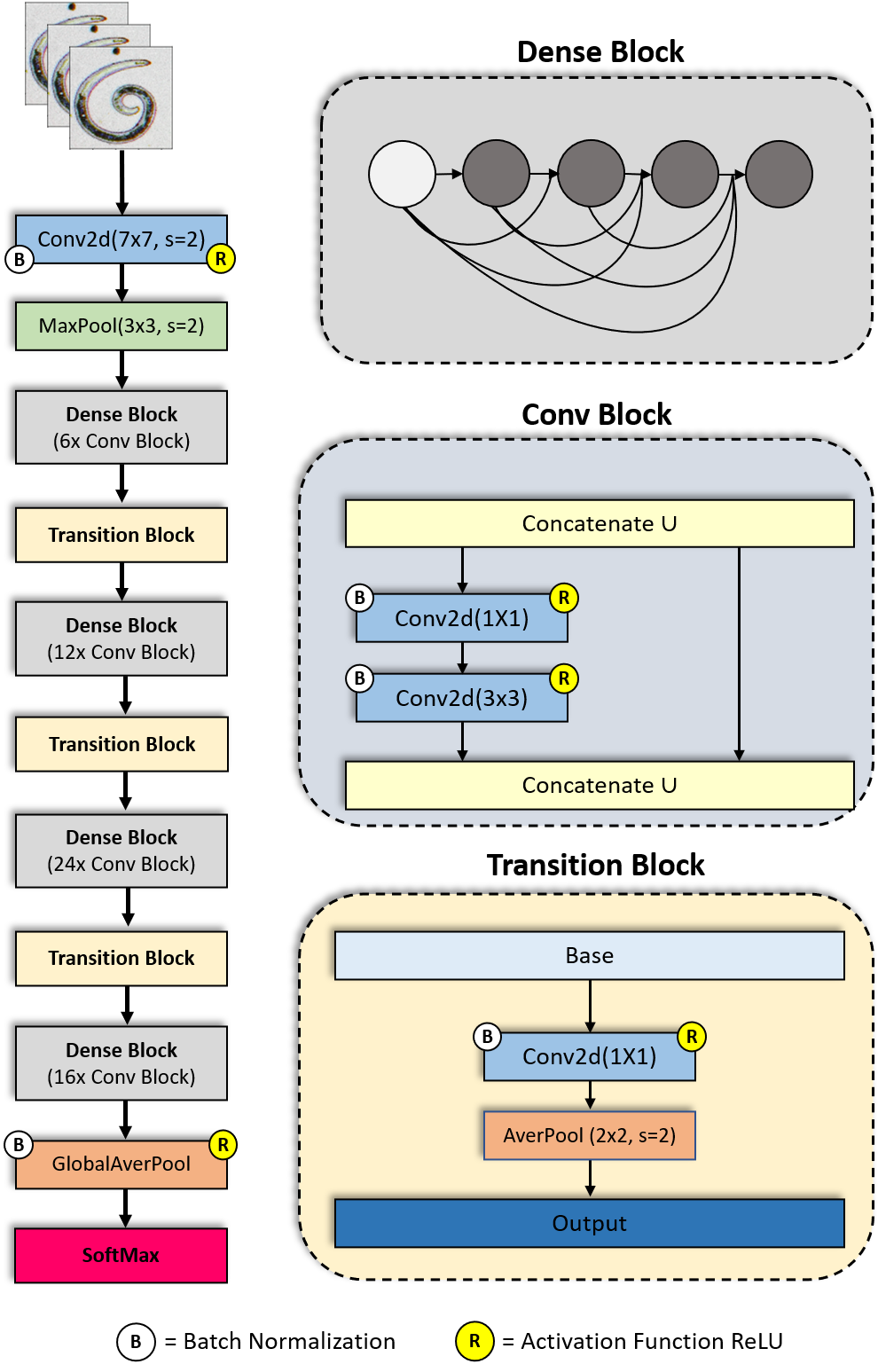}
\caption{\textbf{(Left)} DenseNet121 Architecture. \textbf{(Right)} Dense Block, Conv Block and Transition Block. - Adapted \cite{DenseNet:2017}}\label{fig:Dense121}
\end{figure}

Figure \ref{fig:Dense121} presents the architecture of DenseNet121, which is composed of three types of blocks in its implementation. The first is the convolution block, which is a basic block of a dense block. Convolution block is similar to the identity block in ResNet\cite{ResNet:2016}. The second is a dense block, in which convolution blocks are concatenated and densely connected. The dense block is the main component in DenseNet. The last is the transition layer, which connects two contiguous dense blocks. Since feature map sizes are the same within the dense block, the transition layer reduces the feature map's dimensions.


\begin{figure}[!htpb]
  \centering
  \includegraphics[width=0.7\columnwidth]{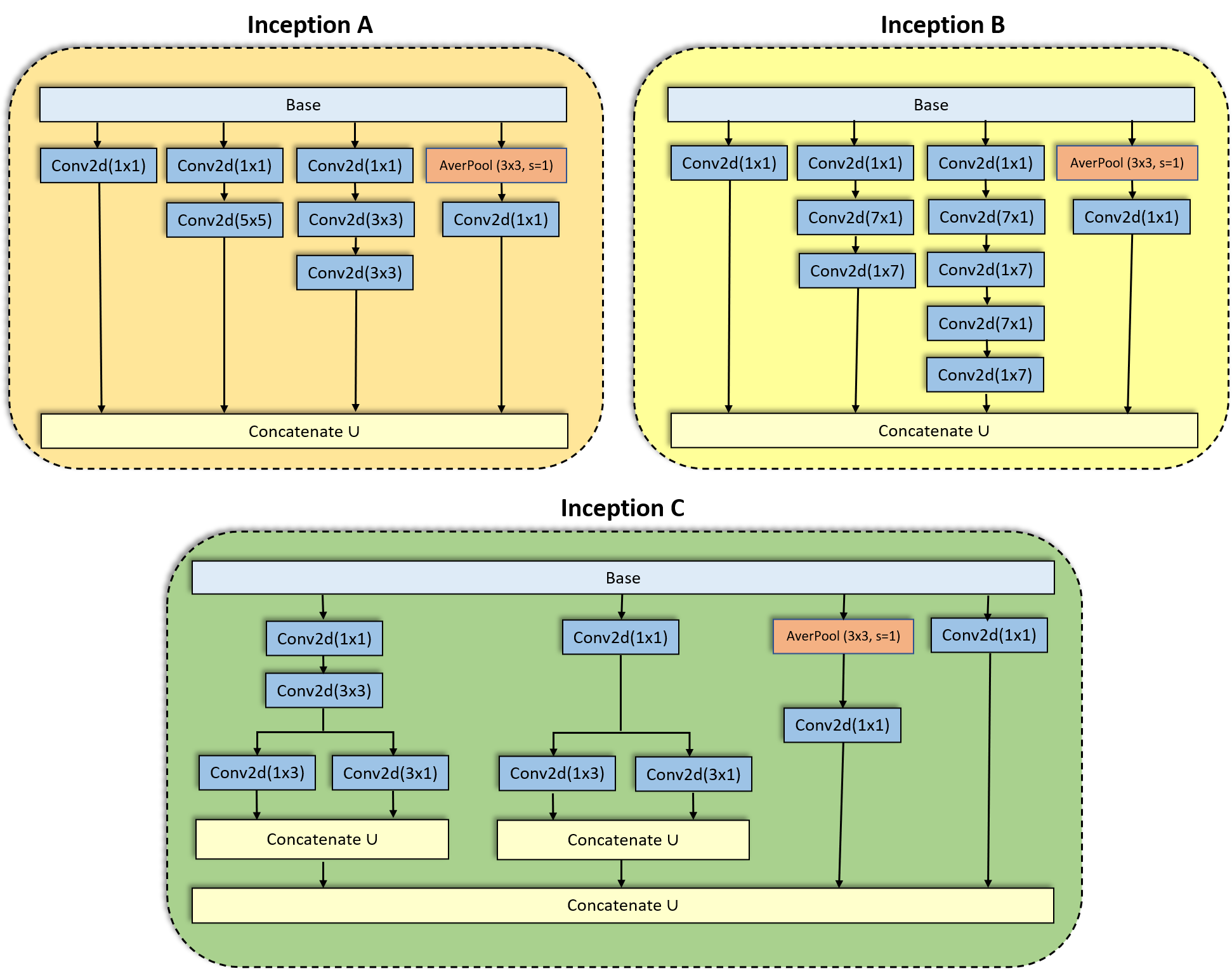}
\caption{Inception Blocks implemented in the InceptionV3 architecture. }\label{fig:InceptionBlocks}
\end{figure}

The Inception Blocks used, on the other hand, implement a multi-scalar approach. Each block has multiple branches with different sizes of kernels ([1 × 1], [3 × 3], [5 × 5], and [7 × 7]). These filters extract and concatenate on a different scale of feature maps and send the combination to the next stage. The 1 × 1 convolution in each inception module is used for dimensionality reduction before applying computationally expensive [3 × 3] and [5 × 5] convolutions. Factorization of [5 × 5], [7 × 7] convolution into smaller convolutions [3 × 3] or asymmetric convolutions ([1 × 7], [7 × 1]) reduces the number of CNN parameters\cite{InceptionV3:2016}. In Figure \ref{fig:InceptionBlocks}, we present the blocks defined as Inception A, B, and C, that were implemented in the NemaNet architecture. 

\begin{figure}[!ht]
  \centering
  \includegraphics[width=0.38\columnwidth]{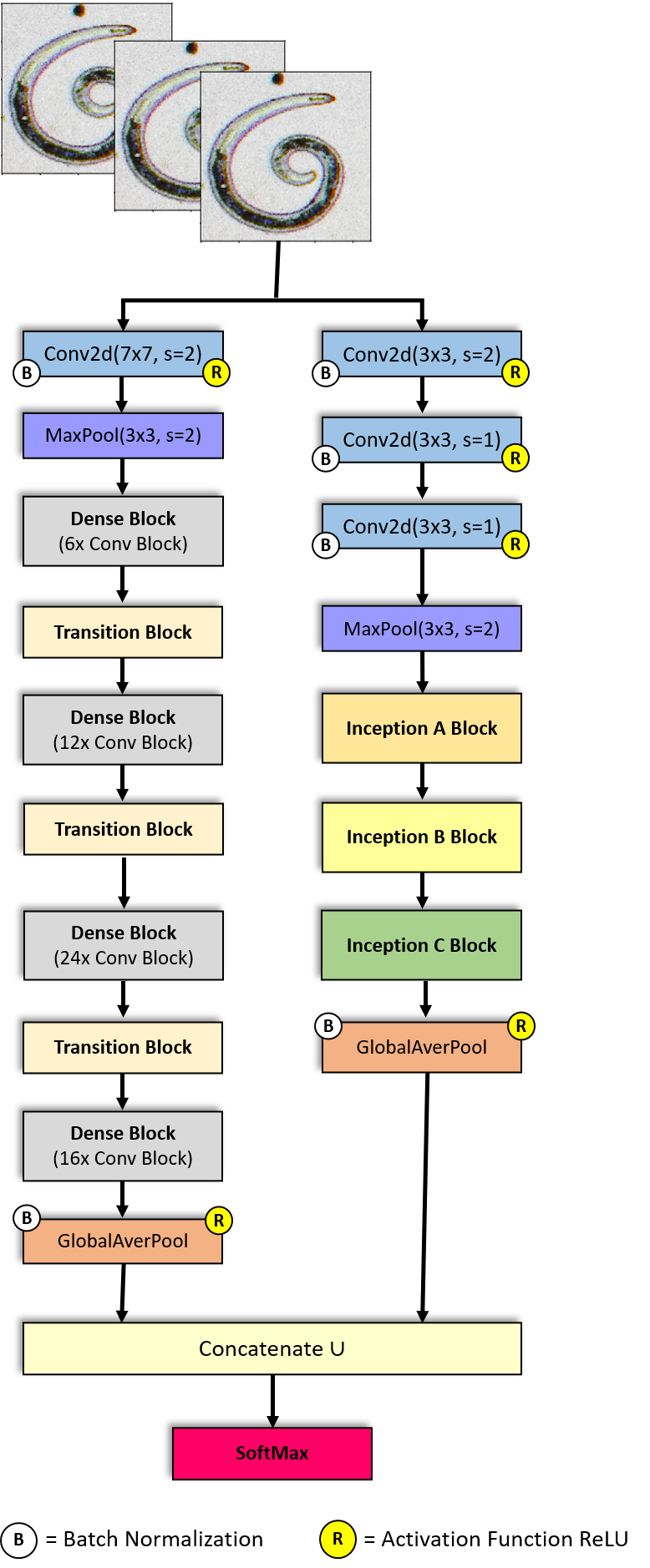}
  \vspace{-0.35cm}
\caption{NemaNet Architecture Cutomized with the structure of DenseNet121 and Inception Blocks}\label{fig:NemaNet}
\end{figure}

After the deep extraction of resources that occurs in parallel by Inception blocks and the structure of DenseNet, each layer obtains additional inputs from all previous layers and passes its feature maps to all subsequent layers. The Concatenation stage is implemented, allowing each layer to share the signal from all previous layers. In Figure \ref{fig:NemaNet} we present our customized architecture proposal called NemaNet CNN.

NemaNet has a total of 17,918,565 parameters, 17,817,925 of which are trainable and 100,640 non-trainable.

\subsection{Dataset Training and Validation}

In this work, we evaluated each architecture using two training approaches: Transfer Learning and From Scratch. From now we describe Transfer Learning as \textbf{TL} and From Scratch as \textbf{FS}. Transfer learning (TL) is a method that reuses models applied to specific tasks as a starting point for a model related to a new domain of interest. Thus, the objective is to borrow labeled or extract knowledge from some related fields to obtain the highest possible performance in the area of interest~\cite{TransferLearning:2010,TransferLearning:2015}.

As per standard practices, there are two ways to approach TL~\cite{Khandelwal:2019}. Using a base neural network as a fixed feature extractor, the images of the target dataset are fed to the deep neural network. The features that generate as input to the final classifier layer are extracted. Through these features, a new classifier is built, and the model is created. In the base network on the last layer of the classifier, the Fine-Tuning is replaced, and the weights of previous layers are also modified. We use pre-trained models on ImageNet~\cite{ImageNet:2009} and implement the transfer of learning using fine-tuning appropriate to the peculiarities of NemaDataset.

Training from scratch (FS) is when the network weights are not inherited from a previous model but are randomly initialized. It requires a larger training set, and the overfitting~\cite{Overfitting:2004} risk is higher since the network has no experience from previous training sessions and so must rely on the input data to define all its weights. However, this approach allows us to define a problem-specific network topology that can improve the performance.

To prevent overfitting due to a limited supply of data and improve the model’s generalization, data augmentations through small random transformations with rotate, flip and mirror, were used on blocks for each epoch. A stochastic optimization algorithm (SGD) was used for optimization to training the proposed network. We initially set a base learning rate as $1\times10^{-3}$. The base learning rate was decreased to $6\times10^{-6}$ with increased iterations.

In the validation process, we used the \textit{k-fold} cross-validation method~\cite{hastie_09,kohavi1995study}. The dataset was divided into 5 ($k$) mutually exclusive subsets of the same size. This strategy causes a subset to be used for the tests and the remaining $k-1$ to estimate the parameters, thus computing the accuracy of the model.

\subsection{Performance Metrics}

To evaluate the proposed architectures' classification performance, the loss, overall accuracy, \textit{F}$_{1}$-scores, precision, recall, and specificity were selected as the accuracy performance metrics. In the training stage, the internal weights of the model are updated during several iterations. We monitor each iteration by training period, saving the weights with the best predictive capacity of the model determined by the validation overall accuracy metric. Table~\ref{Tab:Formulas} below shows how the metrics are calculated.

\begin{table*}[h]
\caption{Metrics used to evaluate the performance of the investigated CNNs architectures}
\label{Tab:Formulas}

	\fontencoding{T1}
    \fontfamily{\sfdefault}
    \fontseries{m}
    \fontshape{n}
    \fontsize{6}{7.5}
    \selectfont
    \setlength{\tabcolsep}{4pt}

\centering
\begin{tabular}{lll}
\hline
Metric &
  Formula &
  Evaluation focus \\ \hline \hline \\
Loss &
  $ \displaystyle L\left ( \hat{y_{i}}, y_{i} \right ) = -\sum_{i=1}^{k} y_{i} \cdot log\left ( \hat{y_{i}}  \right )$ &
  \begin{tabular}[c]{@{}l@{}}A loss function is a method of evaluating how well the model the dataset. \\ The loss function will output a higher number if the predictions are off the actual target  \\ values whereas otherwise it will output a lower number. \\ Since our problem is of type multi-class classification we will be using cross entropy \\ as our loss function.\end{tabular} \\ \\ \hline \\
Accuracy &
  $\displaystyle \sum {_{i=1} ^{k}} \frac{tp_i + tn_i}{tp_i+tn_i+fp_i+tn_i}$ &
  \begin{tabular}[c]{@{}l@{}}The accuracy of a machine learning classification algorithm is one way to measure how  \\often the algorithm classifies a data point correctly.\\ Number of items correctly identified as either truly positive or truly negative out of the total \\ number of items.\end{tabular} \\ \\ \hline \\
\textit{F}$_{1}$-score &
  $\displaystyle 2* \frac{Precision * Recall}{Precision + Recall}$ & \begin{tabular}[c]{@{}l@{}}The harmonic average of the precision and recall, it measures the effectiveness of identification \\ when just as much importance is given to recall as to precision. \end{tabular} \\ \\ \hline \\
Precision &
  $\displaystyle \frac{\sum {_{i=1}^{k} tp_i}}{\sum {_{i=1}^{k}}(tp_i+fp_i)}$ & \begin{tabular}[c]{@{}l@{}} Agreement of the true class labels with those of the classifier's, caculated by summing all TP's\\ and FP's in the system, across all classes. \end{tabular} \\ \\ \hline \\
Recall &
  $\displaystyle \frac{\sum {_{i=1}^{k} tp_i}}{\sum {_{i=1}^{k}}(tp_i+fn_i)}$ & \begin{tabular}[c]{@{}l@{}} Effectiveness of a classifier to identify class labels, calculated by summing all TP's and FN's\\ in the system, across all classes. \end{tabular} \\ \\ \hline \\
Specificity &
  $\displaystyle \frac{\sum {_{i=1}^{k} tn_i}}{\sum {_{i=1}^{k}}(tn_i+fp_i)}$ &\begin{tabular}[c]{@{}l@{}} Specificity is known as the True Negative Rate. This function calculates the proportion of actual\\ negative cases that have gotten predicted as negative by our model.\end{tabular} \\ \\ \hline \hline
\end{tabular}

   {k = total number of classes; tp = true positives; fp = false positives; tn = true negatives; fn = false negatives}
 \\
\end{table*}

\subsubsection{Confusion Matrix}

A confusion matrix~\cite{ConfusionMatrix:2011} contains information about actual and predicted classifications done by a classification system. The performance of the implemented architectures was evaluated using the matrix data, helping to find and eliminate bias and variance problems, enabling adjustments capable of producing more accurate results. The confusion matrix (CM) can be denoted as in Equation~\ref{Equation1}:

    

\begin{equation}\label{Equation1}
\fontsize{6}{6}
\linespread{0.9}
    CM= \begin{bmatrix}
 RR_{1,1}&  RR_{1,2}& \cdots  &RR_{1,N} \\ 
 \vdots &  \vdots &  &  \vdots \\ 
RR_{2,1}&  RR_{2,2}& \cdots  &RR_{2,N} \\  
\vdots &  \vdots &  &  \vdots \\ 
RR_{N,1}&  RR_{N,2}& \cdots  &RR_{N,N}
\end{bmatrix}
\end{equation}

where $RR_{i,j}$ corresponds to the total number of entities in class $C_{i}$ which have been classified in class $C_{j}$. Hence, the main diagonal elements indicate the total number of samples in class $C_{i}$ correctly recognized by the system.

\subsubsection{Receiving Operator Characteristics-Area Under Curve Analysis}


The Receiving Operator Characteristics - Area Under Curve(ROC-AUC)~\cite{ROC:2011} helps to analyze classification performance in various threshold settings. High True Positive Rates (TPR/Sensitivity) of a class describes that the model has performed well in classifying that particular class. ROC-AUC curves can be compared for various models, and the model that possesses a high AUC is considered to have performed well.

\subsection{Software and hardware system Description}

The experiments were implemented on a Linux machine, Ubuntu 18.04, Intel® Core(TM) i7-6800K processor, 2 Nvidia® GTX Titan Xp 12GB GPUs, and 64GB of DDR4 RAM. All models were developed using TensorFlow API version 1.14~\cite{tensorflow} and Keras version 2.2.5~\cite{keras}. For algorithm implementation, and data wrangling was used Python 3.6~\cite{Python:2009}.

\section{Results} \label{sec:results}

This section presents the results obtained based on the CNNs models detailed in Section~\ref{subsec-cnn}. All different CNNs models were trained using the training parameters shown in Table~\ref{Tab:CNNModels}. Each model was evaluated using the FS and TL training approach, implementing cross-validation with five folds ($k=5$). We tried to standardize the Batch Size according to the references of the pre-trained models using ImageNet. However, some models exceeded our computational capacity and were adjusted correctly.

Table~\ref{Tab:Results} presents the overall accuracies obtained for each iteration of the cross-validation using NemaDataset. At the end of each type of training performed, we present an average of the values for each used metric.

\begin{center}
   	\fontencoding{T1}
    \fontfamily{\sfdefault}
    \fontseries{m}
    \fontshape{n}
    \fontsize{5.5}{5.5}
    \selectfont
    \setlength{\tabcolsep}{7pt}
\vspace{-0.5cm}
\begin{longtable}[c]{cllrrrrrr}
\caption{Metric score for different models developed from this study} \label{Tab:Results}
\\

\hline
\multirow{2}{*}{CNN Models} &
  \multicolumn{1}{c}{\multirow{2}{*}{Type Training}} &
  \multicolumn{1}{c}{\multirow{2}{*}{K-Fold 5}} &
  \multicolumn{6}{c}{Metrics} \\ \cline{4-9} 
 &
  \multicolumn{1}{c}{} &
  \multicolumn{1}{c}{} &
  \multicolumn{1}{c}{Loss} &
  \multicolumn{1}{c}{Accuracy} &
  \multicolumn{1}{c}{\textit{F}$_{1}$-score} &
  \multicolumn{1}{c}{Precision} &
  \multicolumn{1}{c}{Recall} &
  \multicolumn{1}{c}{Specificity} \\ \hline
\endfirsthead
\multicolumn{9}{c}%
{{\bfseries Table \thetable\ continued from previous page}} \\
\hline
\multirow{2}{*}{CNN Models} &
  \multicolumn{1}{c}{\multirow{2}{*}{Type Training}} &
  \multicolumn{1}{c}{\multirow{2}{*}{K-Fold 5}} &
  \multicolumn{6}{c}{Metrics} \\ \cline{4-9} 
 &
  \multicolumn{1}{c}{} &
  \multicolumn{1}{c}{} &
  \multicolumn{1}{c}{Loss} &
  \multicolumn{1}{c}{Accuracy} &
  \multicolumn{1}{c}{\textit{F}$_{1}$-score} &
  \multicolumn{1}{c}{Precision} &
  \multicolumn{1}{c}{Recall} &
  \multicolumn{1}{c}{Specificity} \\ \hline
\endhead
\hline
\endfoot
\endlastfoot
\multicolumn{1}{c|}{\multirow{12}{*}{\begin{tabular}[c]{@{}c@{}}Xception \\ 100 Epochs\\ Batch Size = 32\end{tabular}}} &
  \multicolumn{1}{c}{\multirow{5}{*}{FS}} &
  Fold 1 &
  0,1743 &
  0,9592 &
  0,9545 &
  0,9633 &
  0,9462 &
  0,9748 \\
\multicolumn{1}{c|}{} &
  \multicolumn{1}{c}{} &
  Fold 2 &
  0,2565 &
  0,9217 &
  0,9189 &
  0,9262 &
  0,9119 &
  0,9681 \\
\multicolumn{1}{c|}{} &
  \multicolumn{1}{c}{} &
  Fold 3 &
  0,1498 &
  0,9560 &
  0,9542 &
  0,9645 &
  0,9445 &
  0,9749 \\
\multicolumn{1}{c|}{} &
  \multicolumn{1}{c}{} &
  Fold 4 &
  0,1829 &
  0,9412 &
  0,9395 &
  0,9481 &
  0,9314 &
  0,9729 \\
\multicolumn{1}{c|}{} &
  \multicolumn{1}{c}{} &
  Fold 5 &
  0,1584 &
  0,9526 &
  0,9559 &
  0,9697 &
  0,9428 &
  0,9731 \\ \cline{2-9} 
\multicolumn{1}{c|}{} &
  \multicolumn{2}{l}{Metrics Average} &
  0,1844 &
  0,9461 &
  0,9446 &
  0,9544 &
  0,9354 &
  0,9728 \\ \cline{2-9} &
\multicolumn{1}{|c}{\multirow{5}{*}{TL}} &
  Fold 1 &
  0,1276 &
  0,9657 &
  0,9665 &
  0,9673 &
  0,9657 &
  0,9880 \\
\multicolumn{1}{c|}{} &
   &
  Fold 2 &
  0,1451 &
  0,9543 &
  0,9517 &
  0,9540 &
  0,9494 &
  0,9842 \\
\multicolumn{1}{c|}{} &
   &
  Fold 3 &
  0,1908 &
  0,9592 &
  0,9574 &
  0,9588 &
  0,9560 &
  0,9849 \\
\multicolumn{1}{c|}{} &
   &
  Fold 4 &
  0,3148 &
  0,9314 &
  0,9305 &
  0,9313 &
  0,9297 &
  0,9795 \\
\multicolumn{1}{c|}{} &
   &
  Fold 5 &
  0,1719 &
  0,9575 &
  0,9590 &
  0,9606 &
  0,9575 &
  0,9864 \\ \cline{2-9} 
\multicolumn{1}{c|}{} &
  \multicolumn{2}{l}{Metrics Average} &
  0,1900 &
  0,9536 &
  0,9530 &
  0,9544 &
  0,9517 &
  0,9846 \\ \hline
\multicolumn{1}{c|}{\multirow{12}{*}{\begin{tabular}[c]{@{}c@{}}VGG16\\ 100 Epochs\\ Batch Size = 32\end{tabular}}} &
  \multicolumn{1}{c}{\multirow{5}{*}{FS}} &
  Fold 1 &
  1,0559 &
  0,7830 &
  0,7846 &
  0,7931 &
  0,7765 &
  0,9412 \\
\multicolumn{1}{c|}{} &
  \multicolumn{1}{c}{} &
  Fold 2 &
  0,9878 &
  0,7586 &
  0,7569 &
  0,7654 &
  0,7488 &
  0,9338 \\
\multicolumn{1}{c|}{} &
  \multicolumn{1}{c}{} &
  Fold 3 &
  0,8638 &
  0,8189 &
  0,8195 &
  0,8235 &
  0,8157 &
  0,9469 \\
\multicolumn{1}{c|}{} &
  \multicolumn{1}{c}{} &
  Fold 4 &
  1,4808 &
  0,7925 &
  0,7942 &
  0,7960 &
  0,7925 &
  0,9456 \\
\multicolumn{1}{c|}{} &
  \multicolumn{1}{c}{} &
  Fold 5 &
  1,4013 &
  0,8023 &
  0,8035 &
  0,8048 &
  0,8023 &
  0,9486 \\ \cline{2-9} 
\multicolumn{1}{c|}{} &
  \multicolumn{2}{l}{Metrics Average} &
  1,1579 &
  0,7911 &
  0,7917 &
  0,7966 &
  0,7871 &
  0,9432 \\ \cline{2-9} &
\multicolumn{1}{|c}{\multirow{5}{*}{TL}} &
  Fold 1 &
  0,2284 &
  0,9315 &
  0,9314 &
  0,9382 &
  0,9250 &
  0,9745 \\
\multicolumn{1}{c|}{} &
   &
  Fold 2 &
  0,2661 &
  0,9250 &
  0,9249 &
  0,9300 &
  0,9201 &
  0,9733 \\
\multicolumn{1}{c|}{} &
   &
  Fold 3 &
  0,2956 &
  0,9070 &
  0,9072 &
  0,9142 &
  0,9005 &
  0,9704 \\
\multicolumn{1}{c|}{} &
   &
  Fold 4 &
  0,2495 &
  0,9265 &
  0,9257 &
  0,9336 &
  0,9183 &
  0,9721 \\
\multicolumn{1}{c|}{} &
   &
  Fold 5 &
  0,2845 &
  0,9150 &
  0,9128 &
  0,9172 &
  0,9085 &
  0,9684 \\ \cline{2-9} 
\multicolumn{1}{c|}{} &
  \multicolumn{2}{l}{Metrics Average} &
  0,2648 &
  0,9210 &
  0,9204 &
  0,9267 &
  0,9145 &
  0,9718 \\ \hline
\multicolumn{1}{c|}{\multirow{12}{*}{\begin{tabular}[c]{@{}c@{}}InceptionV3 \\ 100 Epochs\\ Batch Size = 32\end{tabular}}} &
  \multicolumn{1}{c}{\multirow{5}{*}{FS}} &
  Fold 1 &
  0,1570 &
  0,9625 &
  0,9620 &
  0,9666 &
  0,9576 &
  0,9860 \\
\multicolumn{1}{c|}{} &
  \multicolumn{1}{c}{} &
  Fold 2 &
  0,1317 &
  0,9592 &
  0,9599 &
  0,9641 &
  0,9560 &
  0,9854 \\
\multicolumn{1}{c|}{} &
  \multicolumn{1}{c}{} &
  Fold 3 &
  0,1697 &
  0,9608 &
  0,9607 &
  0,9623 &
  0,9592 &
  0,9857 \\
\multicolumn{1}{c|}{} &
  \multicolumn{1}{c}{} &
  Fold 4 &
  0,1971 &
  0,9608 &
  0,9615 &
  0,9641 &
  0,9592 &
  0,9858 \\
\multicolumn{1}{c|}{} &
  \multicolumn{1}{c}{} &
  Fold 5 &
  0,1651 &
  0,9493 &
  0,9484 &
  0,9507 &
  0,9461 &
  0,9848 \\ \cline{2-9} 
\multicolumn{1}{c|}{} &
  \multicolumn{2}{l}{Metrics Average} &
  0,1641 &
  0,9585 &
  0,9585 &
  0,9616 &
  0,9556 &
  0,9855 \\ \cline{2-9}&
\multicolumn{1}{|c}{\multirow{5}{*}{TL}} &
  Fold 1 &
  0,2181 &
  0,9445 &
  0,9452 &
  0,9459 &
  0,9445 &
  0,9814 \\
\multicolumn{1}{c|}{} &
   &
  Fold 2 &
  0,1532 &
  0,9592 &
  0,9583 &
  0,9608 &
  0,9560 &
  0,9833 \\
\multicolumn{1}{c|}{} &
   &
  Fold 3 &
  0,1701 &
  0,9641 &
  0,9614 &
  0,9636 &
  0,9592 &
  0,9847 \\
\multicolumn{1}{c|}{} &
   &
  Fold 4 &
  0,1197 &
  0,9657 &
  0,9648 &
  0,9673 &
  0,9624 &
  0,9873 \\
\multicolumn{1}{c|}{} &
   &
  Fold 5 &
  0,1516 &
  0,9657 &
  0,9639 &
  0,9655 &
  0,9624 &
  0,9867 \\ \cline{2-9} 
\multicolumn{1}{c|}{} &
  \multicolumn{2}{l}{Metrics Average} &
  0,1626 &
  0,9598 &
  0,9587 &
  0,9606 &
  0,9569 &
  0,9847 \\ \hline
\multicolumn{1}{c|}{\multirow{12}{*}{\begin{tabular}[c]{@{}c@{}}ResNet50\\ 100 Epochs\\ Batch Size = 100\end{tabular}}} &
  \multicolumn{1}{c}{\multirow{5}{*}{FS}} &
  Fold 1 &
  0,3027 &
  0,9152 &
  0,9136 &
  0,9202 &
  0,9070 &
  0,9745 \\
\multicolumn{1}{c|}{} &
  \multicolumn{1}{c}{} &
  Fold 2 &
  0,3082 &
  0,9119 &
  0,9144 &
  0,9221 &
  0,9070 &
  0,9702 \\
\multicolumn{1}{c|}{} &
  \multicolumn{1}{c}{} &
  Fold 3 &
  0,3014 &
  0,9005 &
  0,8988 &
  0,9106 &
  0,8874 &
  0,9649 \\
\multicolumn{1}{c|}{} &
  \multicolumn{1}{c}{} &
  Fold 4 &
  0,3192 &
  0,9101 &
  0,9071 &
  0,9123 &
  0,9020 &
  0,9718 \\
\multicolumn{1}{c|}{} &
  \multicolumn{1}{c}{} &
  Fold 5 &
  0,3207 &
  0,9069 &
  0,9030 &
  0,9090 &
  0,8971 &
  0,9707 \\ \cline{2-9} 
\multicolumn{1}{c|}{} &
  \multicolumn{2}{l}{Metrics Average} &
  0,3105 &
  0,9089 &
  0,9074 &
  0,9149 &
  0,9001 &
  0,9704 \\ \cline{2-9}&
\multicolumn{1}{|c}{\multirow{5}{*}{TL}} &
  Fold 1 &
  0,1508 &
  0,9592 &
  0,9588 &
  0,9651 &
  0,9527 &
  0,9797 \\
\multicolumn{1}{c|}{} &
   &
  Fold 2 &
  0,1351 &
  0,9462 &
  0,9432 &
  0,9502 &
  0,9364 &
  0,9787 \\
\multicolumn{1}{c|}{} &
   &
  Fold 3 &
  0,1578 &
  0,9445 &
  0,9465 &
  0,9519 &
  0,9413 &
  0,9783 \\
\multicolumn{1}{c|}{} &
   &
  Fold 4 &
  0,2210 &
  0,9281 &
  0,9293 &
  0,9339 &
  0,9248 &
  0,9717 \\
\multicolumn{1}{c|}{} &
   &
  Fold 5 &
  0,1485 &
  0,9641 &
  0,9621 &
  0,9685 &
  0,9559 &
  0,9780 \\ \cline{2-9} 
\multicolumn{1}{c|}{} &
  \multicolumn{2}{l}{Metrics Average} &
  0,1627 &
  0,9484 &
  0,9480 &
  0,9539 &
  0,9422 &
  0,9773 \\ \hline 
\multicolumn{1}{c|}{\multirow{12}{*}{\begin{tabular}[c]{@{}c@{}}ResNet101\\ 100 Epochs\\ Batch Size = 64\end{tabular}}} &
  \multicolumn{1}{c}{\multirow{5}{*}{FS}} &
  Fold 1 &
  0,3906 &
  0,9184 &
  0,9182 &
  0,9197 &
  0,9168 &
  0,9749 \\
\multicolumn{1}{c|}{} &
  \multicolumn{1}{c}{} &
  Fold 2 &
  0,3984 &
  0,9038 &
  0,9048 &
  0,9092 &
  0,9005 &
  0,9708 \\
\multicolumn{1}{c|}{} &
  \multicolumn{1}{c}{} &
  Fold 3 &
  0,4824 &
  0,8842 &
  0,8873 &
  0,8922 &
  0,8825 &
  0,9665 \\
\multicolumn{1}{c|}{} &
  \multicolumn{1}{c}{} &
  Fold 4 &
  0,4026 &
  0,8987 &
  0,9005 &
  0,9041 &
  0,8971 &
  0,9715 \\
\multicolumn{1}{c|}{} &
  \multicolumn{1}{c}{} &
  Fold 5 &
  0,3444 &
  0,9069 &
  0,9075 &
  0,9114 &
  0,9036 &
  0,9721 \\ \cline{2-9} 
\multicolumn{1}{c|}{} &
  \multicolumn{2}{l}{Metrics Average} &
  0,4037 &
  0,9024 &
  0,9037 &
  0,9073 &
  0,9001 &
  0,9712 \\ \cline{2-9}&
\multicolumn{1}{|c}{\multirow{5}{*}{TL}} &
  Fold 1 &
  0,1583 &
  0,9413 &
  0,9400 &
  0,9470 &
  0,9331 &
  0,9786 \\
\multicolumn{1}{c|}{} &
   &
  Fold 2 &
  0,1715 &
  0,9511 &
  0,9465 &
  0,9519 &
  0,9413 &
  0,9805 \\
\multicolumn{1}{c|}{} &
   &
  Fold 3 &
  0,1944 &
  0,9445 &
  0,9449 &
  0,9520 &
  0,9380 &
  0,9793 \\
\multicolumn{1}{c|}{} &
   &
  Fold 4 &
  0,1370 &
  0,9673 &
  0,9663 &
  0,9704 &
  0,9624 &
  0,9818 \\
\multicolumn{1}{c|}{} &
   &
  Fold 5 &
  0,1556 &
  0,9477 &
  0,9483 &
  0,9522 &
  0,9444 &
  0,9814 \\ \cline{2-9} 
\multicolumn{1}{c|}{} &
  \multicolumn{2}{l}{Metrics Average} &
  0,1634 &
  0,9504 &
  0,9492 &
  0,9547 &
  0,9439 &
  0,9803 \\ \hline 
\multicolumn{1}{c|}{\multirow{12}{*}{\begin{tabular}[c]{@{}c@{}}ResNet152\\ 100 Epochs\\ Batch Size = 64\end{tabular}}} &
  \multicolumn{1}{c}{\multirow{5}{*}{FS}} &
  Fold 1 &
  0,4002 &
  0,8809 &
  0,8825 &
  0,8891 &
  0,8760 &
  0,9620 \\
\multicolumn{1}{c|}{} &
  \multicolumn{1}{c}{} &
  Fold 2 &
  0,2761 &
  0,9103 &
  0,9074 &
  0,9111 &
  0,9038 &
  0,9692 \\
\multicolumn{1}{c|}{} &
  \multicolumn{1}{c}{} &
  Fold 3 &
  0,4031 &
  0,8907 &
  0,8906 &
  0,8956 &
  0,8858 &
  0,9681 \\
\multicolumn{1}{c|}{} &
  \multicolumn{1}{c}{} &
  Fold 4 &
  0,3335 &
  0,9069 &
  0,9087 &
  0,9123 &
  0,9052 &
  0,9721 \\
\multicolumn{1}{c|}{} &
  \multicolumn{1}{c}{} &
  Fold 5 &
  0,3349 &
  0,9150 &
  0,9154 &
  0,9191 &
  0,9118 &
  0,9735 \\ \cline{2-9} 
\multicolumn{1}{c|}{} &
  \multicolumn{2}{l}{Metrics Average} &
  0,3496 &
  0,9008 &
  0,9009 &
  0,9054 &
  0,8965 &
  0,9690 \\ \cline{2-9} &
\multicolumn{1}{|c}{\multirow{5}{*}{TL}} &
  Fold 1 &
  0,1606 &
  0,9462 &
  0,9452 &
  0,9579 &
  0,9331 &
  0,9777 \\
\multicolumn{1}{c|}{} &
   &
  Fold 2 &
  0,1554 &
  0,9592 &
  0,9581 &
  0,9637 &
  0,9527 &
  0,9807 \\
\multicolumn{1}{c|}{} &
   &
  Fold 3 &
  0,1493 &
  0,9543 &
  0,9541 &
  0,9571 &
  0,9511 &
  0,9794 \\
\multicolumn{1}{c|}{} &
   &
  Fold 4 &
  0,1530 &
  0,9575 &
  0,9581 &
  0,9621 &
  0,9542 &
  0,9821 \\
\multicolumn{1}{c|}{} &
   &
  Fold 5 &
  0,2301 &
  0,9346 &
  0,9325 &
  0,9387 &
  0,9265 &
  0,9730 \\ \cline{2-9} 
\multicolumn{1}{c|}{} &
  \multicolumn{2}{l}{Metrics Average} &
  0,1697 &
  0,9504 &
  0,9496 &
  0,9559 &
  0,9435 &
  0,9786 \\ \hline \\
\multicolumn{1}{c|}{\multirow{12}{*}{\begin{tabular}[c]{@{}c@{}}InceptionResNetV2\\ 100 Epochs\\ Batch Size = 32\end{tabular}}} &
  \multicolumn{1}{c}{\multirow{5}{*}{FS}} &
  Fold 1 &
  0,2894 &
  0,9315 &
  0,9283 &
  0,9354 &
  0,9217 &
  0,9770 \\
\multicolumn{1}{c|}{} &
  \multicolumn{1}{c}{} &
  Fold 2 &
  0,1890 &
  0,9380 &
  0,9391 &
  0,9436 &
  0,9347 &
  0,9810 \\
\multicolumn{1}{c|}{} &
  \multicolumn{1}{c}{} &
  Fold 3 &
  0,2531 &
  0,9445 &
  0,9434 &
  0,9490 &
  0,9380 &
  0,9806 \\
\multicolumn{1}{c|}{} &
  \multicolumn{1}{c}{} &
  Fold 4 &
  0,1837 &
  0,9493 &
  0,9471 &
  0,9516 &
  0,9428 &
  0,9814 \\
\multicolumn{1}{c|}{} &
  \multicolumn{1}{c}{} &
  Fold 5 &
  0,2417 &
  0,9346 &
  0,9324 &
  0,9386 &
  0,9265 &
  0,9762 \\ \cline{2-9} 
\multicolumn{1}{c|}{} &
  \multicolumn{2}{l}{Metrics Average} &
  0,2314 &
  0,9396 &
  0,9381 &
  0,9436 &
  0,9327 &
  0,9793 \\ \cline{2-9} &
\multicolumn{1}{|c}{\multirow{5}{*}{TL}} &
  Fold 1 &
  0,1391 &
  0,9576 &
  0,9573 &
  0,9620 &
  0,9527 &
  0,9829 \\
\multicolumn{1}{c|}{} &
   &
  Fold 2 &
  0,1181 &
  0,9608 &
  0,9623 &
  0,9655 &
  0,9592 &
  0,9858 \\
\multicolumn{1}{c|}{} &
   &
  Fold 3 &
  0,0771 &
  0,9723 &
  0,9713 &
  0,9753 &
  0,9674 &
  0,9893 \\
\multicolumn{1}{c|}{} &
   &
  Fold 4 &
  0,2102 &
  0,9477 &
  0,9466 &
  0,9489 &
  0,9444 &
  0,9831 \\
\multicolumn{1}{c|}{} &
   &
  Fold 5 &
  0,1718 &
  0,9624 &
  0,9616 &
  0,9624 &
  0,9608 &
  0,9841 \\ \cline{2-9} 
\multicolumn{1}{c|}{} &
  \multicolumn{2}{l}{Metrics Average} &
  0,1433 &
  0,9602 &
  0,9598 &
  0,9628 &
  0,9569 &
  0,9851 \\ \hline  
\multicolumn{1}{c|}{\multirow{12}{*}{\begin{tabular}[c]{@{}c@{}}DenseNet121\\ 100 Epochs\\ Batch Size = 32\end{tabular}}} &
  \multicolumn{1}{c}{\multirow{5}{*}{FS}} &
  Fold 1 &
  0,2465 &
  0,9201 &
  0,9202 &
  0,9255 &
  0,9152 &
  0,9745 \\
\multicolumn{1}{c|}{} &
  \multicolumn{1}{c}{} &
  Fold 2 &
  0,2963 &
  0,8940 &
  0,8875 &
  0,9043 &
  0,8728 &
  0,9642 \\
\multicolumn{1}{c|}{} &
  \multicolumn{1}{c}{} &
  Fold 3 &
  0,2037 &
  0,9380 &
  0,9359 &
  0,9388 &
  0,9331 &
  0,9801 \\
\multicolumn{1}{c|}{} &
  \multicolumn{1}{c}{} &
  Fold 4 &
  0,1785 &
  0,9461 &
  0,9482 &
  0,9503 &
  0,9461 &
  0,9807 \\
\multicolumn{1}{c|}{} &
  \multicolumn{1}{c}{} &
  Fold 5 &
  0,3485 &
  0,8971 &
  0,9000 &
  0,9030 &
  0,8971 &
  0,9698 \\ \cline{2-9} 
\multicolumn{1}{c|}{} &
  \multicolumn{2}{l}{Metrics Average} &
  0,2547 &
  0,9190 &
  0,9184 &
  0,9244 &
  0,9128 &
  0,9739 \\ \cline{2-9} &
\multicolumn{1}{|c}{\multirow{5}{*}{TL}} &
  Fold 1 &
  0,1091 &
  0,9723 &
  0,9723 &
  0,9723 &
  0,9723 &
  0,9887 \\
\multicolumn{1}{c|}{} &
   &
  Fold 2 &
  0,1225 &
  0,9641 &
  0,9646 &
  0,9703 &
  0,9592 &
  0,9888 \\
\multicolumn{1}{c|}{} &
   &
  Fold 3 &
  0,0956 &
  0,9772 &
  0,9777 &
  0,9800 &
  0,9755 &
  0,9915 \\
\multicolumn{1}{c|}{} &
   &
  Fold 4 &
  0,1145 &
  0,9706 &
  0,9705 &
  0,9721 &
  0,9690 &
  0,9897 \\
\multicolumn{1}{c|}{} &
   &
  Fold 5 &
  0,0523 &
  0,9886 &
  0,9886 &
  0,9886 &
  0,9886 &
  0,9935 \\ \cline{2-9} 
\multicolumn{1}{c|}{} &
  \multicolumn{2}{l}{Metrics Average} &
  0,0988 &
  0,9745 &
  0,9747 &
  0,9766 &
  0,9729 &
  0,9904 \\ \hline  \multicolumn{1}{c|}{\multirow{12}{*}{\begin{tabular}[c]{@{}c@{}}DenseNet169\\ 100 Epochs \\ Batch Size = 32\end{tabular}}} &
  \multicolumn{1}{c}{\multirow{5}{*}{FS}} &
  Fold 1 &
  0,2504 &
  0,9184 &
  0,9234 &
  0,9377 &
  0,9103 &
  0,9695 \\
\multicolumn{1}{c|}{} &
  \multicolumn{1}{c}{} &
  Fold 2 &
  0,2081 &
  0,9429 &
  0,9434 &
  0,9473 &
  0,9396 &
  0,9823 \\
\multicolumn{1}{c|}{} &
  \multicolumn{1}{c}{} &
  Fold 3 &
  0,2437 &
  0,9168 &
  0,9220 &
  0,9328 &
  0,9119 &
  0,9705 \\
\multicolumn{1}{c|}{} &
  \multicolumn{1}{c}{} &
  Fold 4 &
  0,1299 &
  0,9641 &
  0,9653 &
  0,9684 &
  0,9624 &
  0,9849 \\
\multicolumn{1}{c|}{} &
  \multicolumn{1}{c}{} &
  Fold 5 &
  0,2454 &
  0,9265 &
  0,9233 &
  0,9287 &
  0,9183 &
  0,9756 \\ \cline{2-9} 
\multicolumn{1}{c|}{} &
  \multicolumn{2}{l}{Metrics Average} &
  0,2155 &
  0,9337 &
  0,9355 &
  0,9430 &
  0,9285 &
  0,9765 \\ \cline{2-9} &
\multicolumn{1}{|c}{\multirow{5}{*}{TL}} &
  Fold 1 &
  0,0829 &
  0,9804 &
  0,9787 &
  0,9803 &
  0,9772 &
  0,9921 \\
\multicolumn{1}{c|}{} &
   &
  Fold 2 &
  0,1022 &
  0,9772 &
  0,9771 &
  0,9786 &
  0,9755 &
  0,9914 \\
\multicolumn{1}{c|}{} &
   &
  Fold 3 &
  0,0713 &
  0,9821 &
  0,9820 &
  0,9835 &
  0,9804 &
  0,9917 \\
\multicolumn{1}{c|}{} &
   &
  Fold 4 &
  0,0714 &
  0,9739 &
  0,9746 &
  0,9770 &
  0,9722 &
  0,9914 \\
\multicolumn{1}{c|}{} &
   &
  Fold 5 &
  0,0839 &
  0,9755 &
  0,9738 &
  0,9754 &
  0,9722 &
  0,9905 \\ \cline{2-9} 
\multicolumn{1}{c|}{} &
  \multicolumn{2}{l}{Metrics Average} &
  0,0823 &
  0,9778 &
  0,9772 &
  0,9790 &
  0,9755 &
  0,9914 \\ \hline 
\multicolumn{1}{c|}{\multirow{12}{*}{\begin{tabular}[c]{@{}c@{}}DenseNet201\\ 100 Epochs \\ Batch Size = 32\end{tabular}}} &
  \multicolumn{1}{c}{\multirow{5}{*}{FS}} &
  Fold 1 &
  0,4473 &
  0,8923 &
  0,8897 &
  0,8938 &
  0,8858 &
  0,9688 \\
\multicolumn{1}{c|}{} &
  \multicolumn{1}{c}{} &
  Fold 2 &
  0,3460 &
  0,8989 &
  0,8998 &
  0,9026 &
  0,8972 &
  0,9732 \\
\multicolumn{1}{c|}{} &
  \multicolumn{1}{c}{} &
  Fold 3 &
  0,2161 &
  0,9282 &
  0,9283 &
  0,9336 &
  0,9233 &
  0,9801 \\
\multicolumn{1}{c|}{} &
  \multicolumn{1}{c}{} &
  Fold 4 &
  0,3001 &
  0,8987 &
  0,8944 &
  0,9020 &
  0,8873 &
  0,9675 \\
\multicolumn{1}{c|}{} &
  \multicolumn{1}{c}{} &
  Fold 5 &
  0,2187 &
  0,9379 &
  0,9373 &
  0,9435 &
  0,9314 &
  0,9775 \\ \cline{2-9} 
\multicolumn{1}{c|}{} &
  \multicolumn{2}{l}{Metrics Average} &
  0,3056 &
  0,9112 &
  0,9099 &
  0,9151 &
  0,9050 &
  0,9734 \\ \cline{2-9} &
\multicolumn{1}{|c}{\multirow{5}{*}{TL}} &
  Fold 1 &
  0,0964 &
  0,9739 &
  0,9730 &
  0,9738 &
  0,9723 &
  0,9910 \\
\multicolumn{1}{c|}{} &
   &
  Fold 2 &
  0,0975 &
  0,9821 &
  0,9770 &
  0,9818 &
  0,9723 &
  0,9891 \\
\multicolumn{1}{c|}{} &
   &
  Fold 3 &
  0,0719 &
  0,9853 &
  0,9853 &
  0,9853 &
  0,9853 &
  0,9935 \\
\multicolumn{1}{c|}{} &
   &
  Fold 4 &
  0,0895 &
  0,9788 &
  0,9788 &
  0,9788 &
  0,9788 &
  0,9924 \\
\multicolumn{1}{c|}{} &
   &
  Fold 5 &
  0,1182 &
  0,9739 &
  0,9755 &
  0,9771 &
  0,9739 &
  0,9911 \\ \cline{2-9} 
\multicolumn{1}{c|}{} &
  \multicolumn{2}{l}{Metrics Average} &
  0,0947 &
  0,9788 &
  0,9779 &
  0,9794 &
  0,9765 &
  0,9914 \\ \hline
\multicolumn{1}{c|}{\multirow{12}{*}{\begin{tabular}[c]{@{}c@{}}EfficientNetB0\\ 100 Epochs \\ Batch Size = 100\end{tabular}}} &
  \multicolumn{1}{c}{\multirow{5}{*}{FS}} &
  Fold 1 &
  0,4057 &
  0,8891 &
  0,8845 &
  0,8933 &
  0,8760 &
  0,9628 \\
\multicolumn{1}{c|}{} &
  \multicolumn{1}{c}{} &
  Fold 2 &
  0,5144 &
  0,8662 &
  0,8670 &
  0,8727 &
  0,8613 &
  0,9609 \\
\multicolumn{1}{c|}{} &
  \multicolumn{1}{c}{} &
  Fold 3 &
  0,5290 &
  0,8483 &
  0,8484 &
  0,8586 &
  0,8385 &
  0,9501 \\
\multicolumn{1}{c|}{} &
  \multicolumn{1}{c}{} &
  Fold 4 &
  0,5157 &
  0,8546 &
  0,8518 &
  0,8623 &
  0,8415 &
  0,9527 \\
\multicolumn{1}{c|}{} &
  \multicolumn{1}{c}{} &
  Fold 5 &
  0,6678 &
  0,8235 &
  0,8243 &
  0,8317 &
  0,8170 &
  0,9500 \\ \cline{2-9} 
\multicolumn{1}{c|}{} &
  \multicolumn{2}{l}{Metrics Average} &
  0,5265 &
  0,8563 &
  0,8552 &
  0,8637 &
  0,8469 &
  0,9553 \\ \cline{2-9} &
\multicolumn{1}{|c}{\multirow{5}{*}{TL}} &
  Fold 1 &
  0,2507 &
  0,9217 &
  0,9204 &
  0,9273 &
  0,9135 &
  0,9685 \\
\multicolumn{1}{c|}{} &
   &
  Fold 2 &
  0,2208 &
  0,9152 &
  0,9107 &
  0,9212 &
  0,9005 &
  0,9690 \\
\multicolumn{1}{c|}{} &
   &
  Fold 3 &
  0,2304 &
  0,9413 &
  0,9416 &
  0,9486 &
  0,9347 &
  0,9709 \\
\multicolumn{1}{c|}{} &
   &
  Fold 4 &
  0,1762 &
  0,9510 &
  0,9505 &
  0,9584 &
  0,9428 &
  0,9747 \\
\multicolumn{1}{c|}{} &
   &
  Fold 5 &
  0,1929 &
  0,9395 &
  0,9380 &
  0,9449 &
  0,9314 &
  0,9735 \\ \cline{2-9} 
\multicolumn{1}{c|}{} &
  \multicolumn{2}{l}{Metrics Average} &
  0,2142 &
  0,9337 &
  0,9322 &
  0,9401 &
  0,9246 &
  0,9713 \\ \hline
\multicolumn{1}{c|}{\multirow{12}{*}{\begin{tabular}[c]{@{}c@{}}EfficientNetB3\\ 100 Epochs\\ Batch Size = 32\end{tabular}}} &
  \multicolumn{1}{c}{\multirow{5}{*}{FS}} &
  Fold 1 &
  0,3977 &
  0,8923 &
  0,8809 &
  0,9062 &
  0,8581 &
  0,9526 \\
\multicolumn{1}{c|}{} &
  \multicolumn{1}{c}{} &
  Fold 2 &
  0,4080 &
  0,9184 &
  0,9136 &
  0,9259 &
  0,9021 &
  0,9667 \\
\multicolumn{1}{c|}{} &
  \multicolumn{1}{c}{} &
  Fold 3 &
  0,3695 &
  0,8989 &
  0,8959 &
  0,9014 &
  0,8907 &
  0,9705 \\
\multicolumn{1}{c|}{} &
  \multicolumn{1}{c}{} &
  Fold 4 &
  0,3804 &
  0,9052 &
  0,9008 &
  0,9120 &
  0,8905 &
  0,9642 \\
\multicolumn{1}{c|}{} &
  \multicolumn{1}{c}{} &
  Fold 5 &
  0,3239 &
  0,8971 &
  0,8924 &
  0,9049 &
  0,8805 &
  0,9597 \\ \cline{2-9} 
\multicolumn{1}{c|}{} &
  \multicolumn{2}{l}{Metrics Average} &
  0,3759 &
  0,9024 &
  0,8967 &
  0,9101 &
  0,8844 &
  0,9628 \\ \cline{2-9} &
\multicolumn{1}{|c}{\multirow{5}{*}{TL}} &
  Fold 1 &
  0,1427 &
  0,9625 &
  0,9611 &
  0,9668 &
  0,9560 &
  0,9816 \\
\multicolumn{1}{c|}{} &
   &
  Fold 2 &
  0,1417 &
  0,9527 &
  0,9529 &
  0,9600 &
  0,9462 &
  0,9794 \\
\multicolumn{1}{c|}{} &
   &
  Fold 3 &
  0,1350 &
  0,9608 &
  0,9592 &
  0,9608 &
  0,9576 &
  0,9803 \\
\multicolumn{1}{c|}{} &
   &
  Fold 4 &
  0,1950 &
  0,9493 &
  0,9479 &
  0,9550 &
  0,9412 &
  0,9757 \\
\multicolumn{1}{c|}{} &
   &
  Fold 5 &
  0,1734 &
  0,9444 &
  0,9435 &
  0,9459 &
  0,9412 &
  0,9760 \\ \cline{2-9} 
\multicolumn{1}{c|}{} &
  \multicolumn{2}{l}{Metrics Average} &
  0,1576 &
  0,9540 &
  0,9529 &
  0,9577 &
  0,9484 &
  0,9786 \\ \hline \\ \\ 
\multicolumn{1}{c|}{\multirow{12}{*}{\begin{tabular}[c]{@{}c@{}}NASNetLarge\\ 100 Epochs\\ Batch Size = 16\end{tabular}}} &
  \multicolumn{1}{c}{\multirow{5}{*}{FS}} &
  Fold 1 &
  0,2692 &
  0,9462 &
  0,9475 &
  0,9490 &
  0,9462 &
  0,9851 \\
\multicolumn{1}{c|}{} &
  \multicolumn{1}{c}{} &
  Fold 2 &
  0,4203 &
  0,9135 &
  0,9127 &
  0,9135 &
  0,9119 &
  0,9780 \\
\multicolumn{1}{c|}{} &
  \multicolumn{1}{c}{} &
  Fold 3 &
  0,3227 &
  0,9282 &
  0,9282 &
  0,9282 &
  0,9282 &
  0,9792 \\
\multicolumn{1}{c|}{} &
  \multicolumn{1}{c}{} &
  Fold 4 &
  0,2317 &
  0,9395 &
  0,9387 &
  0,9395 &
  0,9379 &
  0,9812 \\
\multicolumn{1}{c|}{} &
  \multicolumn{1}{c}{} &
  Fold 5 &
  0,2908 &
  0,9379 &
  0,9371 &
  0,9432 &
  0,9314 &
  0,9782 \\ \cline{2-9} 
\multicolumn{1}{c|}{} &
  \multicolumn{2}{l}{Metrics Average} &
  0,3069 &
  0,9331 &
  0,9329 &
  0,9347 &
  0,9311 &
  0,9803 \\ \cline{2-9} &
\multicolumn{1}{|c}{\multirow{5}{*}{TL}} &
  Fold 1 &
  0,2122 &
  0,9445 &
  0,9457 &
  0,9505 &
  0,9413 &
  0,9784 \\
\multicolumn{1}{c|}{} &
   &
  Fold 2 &
  0,2298 &
  0,9429 &
  0,9392 &
  0,9439 &
  0,9347 &
  0,9788 \\
\multicolumn{1}{c|}{} &
   &
  Fold 3 &
  0,1443 &
  0,9674 &
  0,9673 &
  0,9690 &
  0,9657 &
  0,9850 \\
\multicolumn{1}{c|}{} &
   &
  Fold 4 &
  0,1664 &
  0,9493 &
  0,9475 &
  0,9508 &
  0,9444 &
  0,9830 \\
\multicolumn{1}{c|}{} &
   &
  Fold 5 &
  0,1944 &
  0,9412 &
  0,9397 &
  0,9452 &
  0,9346 &
  0,9808 \\ \cline{2-9} 
\multicolumn{1}{c|}{} &
  \multicolumn{2}{l}{Metrics Average} &
  0,1894 &
  0,9491 &
  0,9479 &
  0,9519 &
  0,9442 &
  0,9812 \\ \hline 
  
  \multicolumn{1}{c|}{\multirow{12}{*}{\begin{tabular}[c]{@{}c@{}}NemaNet\\ 100 Epochs\\ Batch Size = 16\end{tabular}}} &
  \multicolumn{1}{c}{\multirow{5}{*}{FS}} &
  Fold 1 &
  0,2624 &
  0,9608 &
  0,9564 &
  0,9578 &
  0,9551 &
  0,9878 \\
\multicolumn{1}{c|}{} &
  \multicolumn{1}{c}{} &
  Fold 2 &
  0,4152 &
  0,9673 &
  0,9686 &
  0,9695 &
  0,9679 &
  0,9910 \\
\multicolumn{1}{c|}{} &
  \multicolumn{1}{c}{} &
  Fold 3 &
  0,1546 &
  0,9722 &
  0,9727 &
  0,9727 &
  0,9727 &
  0,9919 \\
\multicolumn{1}{c|}{} &
  \multicolumn{1}{c}{} &
  Fold 4 &
  0,0815 &
  0,9803 &
  0,9806 &
  0,9823 &
  0,9791 &
  0,9939 \\
\multicolumn{1}{c|}{} &
  \multicolumn{1}{c}{} &
  Fold 5 &
  0,1608 &
  0,9689 &
  0,9702 &
  0,9711 &
  0,9695 &
  0,9920 \\ \cline{2-9} 
\multicolumn{1}{c|}{} &
  \multicolumn{2}{l}{Metrics Average} &
  0,2149 &
  0,9699 &
  0,9697 &
  0,9707 &
  0,9689 &
  0,9913 \\ \cline{2-9} &
\multicolumn{1}{|c}{\multirow{5}{*}{TL}} &
  Fold 1 &
  0,0366 &
  0,9918 &
  0,9919 &
  0,9919 &
  0,9919 &
  0,9974 \\
\multicolumn{1}{c|}{} &
   &
  Fold 2 &
  0,0442 &
  0,9902 &
  0,9903 &
  0,9903 &
  0,9903 &
  0,9968 \\
\multicolumn{1}{c|}{} &
   &
  Fold 3 &
  0,3102 &
  0,9853 &
  0,9861 &
  0,9867 &
  0,9855 &
  0,9954 \\
\multicolumn{1}{c|}{} &
   &
  Fold 4 &
  0,0832 &
  0,9836 &
  0,9847 &
  0,9855 &
  0,9839 &
  0,9955 \\
\multicolumn{1}{c|}{} &
   &
  Fold 5 &
  0,0462 &
  0,9934 &
  0,9927 &
  0,9935 &
  0,9919 &
  0,9973 \\ \cline{2-9} 
\multicolumn{1}{c|}{} &
  \multicolumn{2}{l}{Metrics Average} &
  0,1041 &
  0,9888 &
  0,9891 &
  0,9896 &
  0,9887 &
  0,9965 \\  \hline \hline
  
\end{longtable}
\vspace{-0.75cm}
\end{center}

Among the thirteen traditional models and our proposal of customized model called NemaNet, evaluated in our experiments, we highlight the five best models for each type of training with the highest accuracy calculated in the general average.

For the trained models FS, the following stood out in order of relevance: NemaNet (96.99\%); InceptionV3 (95.85\%); Xception (94.61\%); InceptionResNetV2 (93.96\%) and DenseNet169 (93.37\%).

When we implemented the training using TL, the first five models highlighted were: NemaNet (98.88\%); DenseNet201 (97.87\%); DenseNet169 (97.77\%); DenseNet121 (97.45\%) and InceptionResNetV2 (96.01\%).

Our experiments also demonstrated that the proposed custom architecture called NemaNet obtained the best performance between the two training strategies used. We also highlight that our results using TL used the pre-trained weights of ImageNet with a DenseNet121 architecture, taking advantage of the similarity of our customized model with the dense blocks. The part made up of inception blocks was not initialized with pre-trained weights.

\begin{figure}[!ht]
  \centering
  \includegraphics[width=12.5cm]{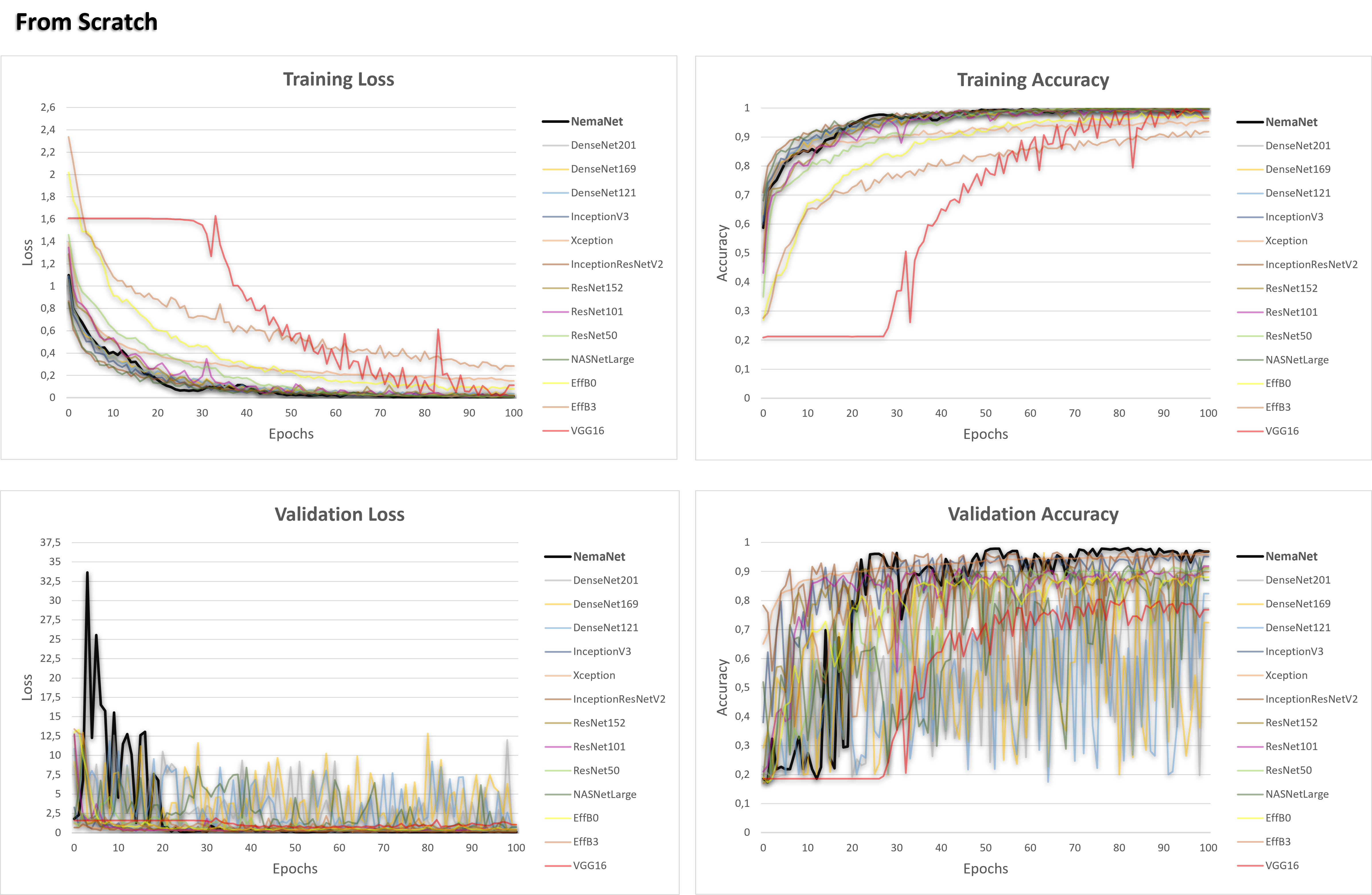}
\caption{The loss and accuracy curves of training process - From Scratch. }\label{fig:FS_Loss_Val}
\end{figure}

\begin{figure}[!ht]
  \centering
  \includegraphics[width=12.5cm]{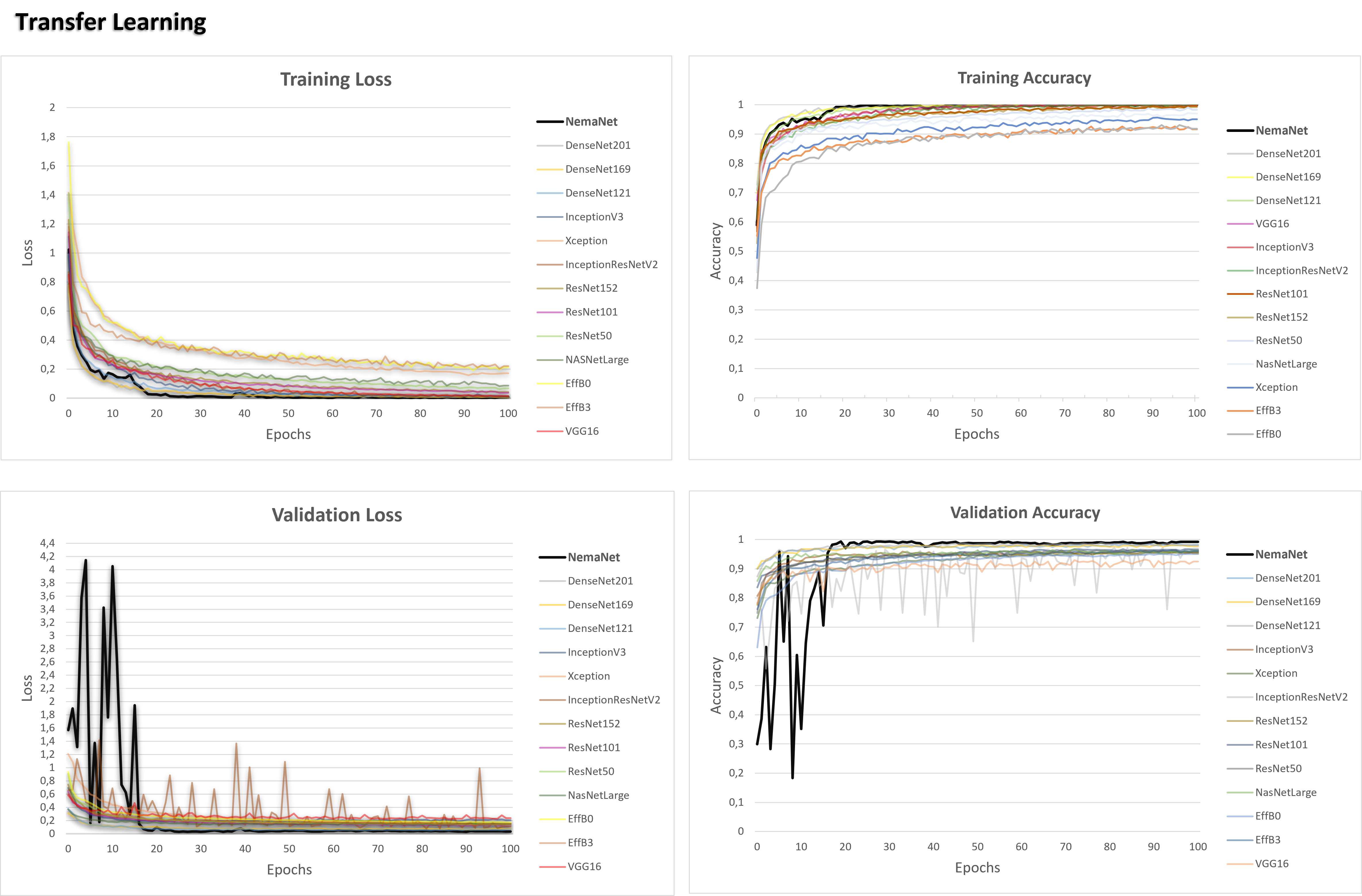}
\caption{The loss and accuracy curves of training process - Transfer Learning.}\label{fig:TL_Loss_Val}
\end{figure}

It was possible to notice that NemaNet, for both FS and TL, started to converge progressively from the twentieth epoch, showing a better behavior among the other evaluated models. Figures~\ref{fig:FS_Loss_Val} and \ref{fig:TL_Loss_Val} we presented the evolution of the training and validation stages of these results.

\begin{figure}[!ht]
  \centering
  \includegraphics[width=13.5cm]{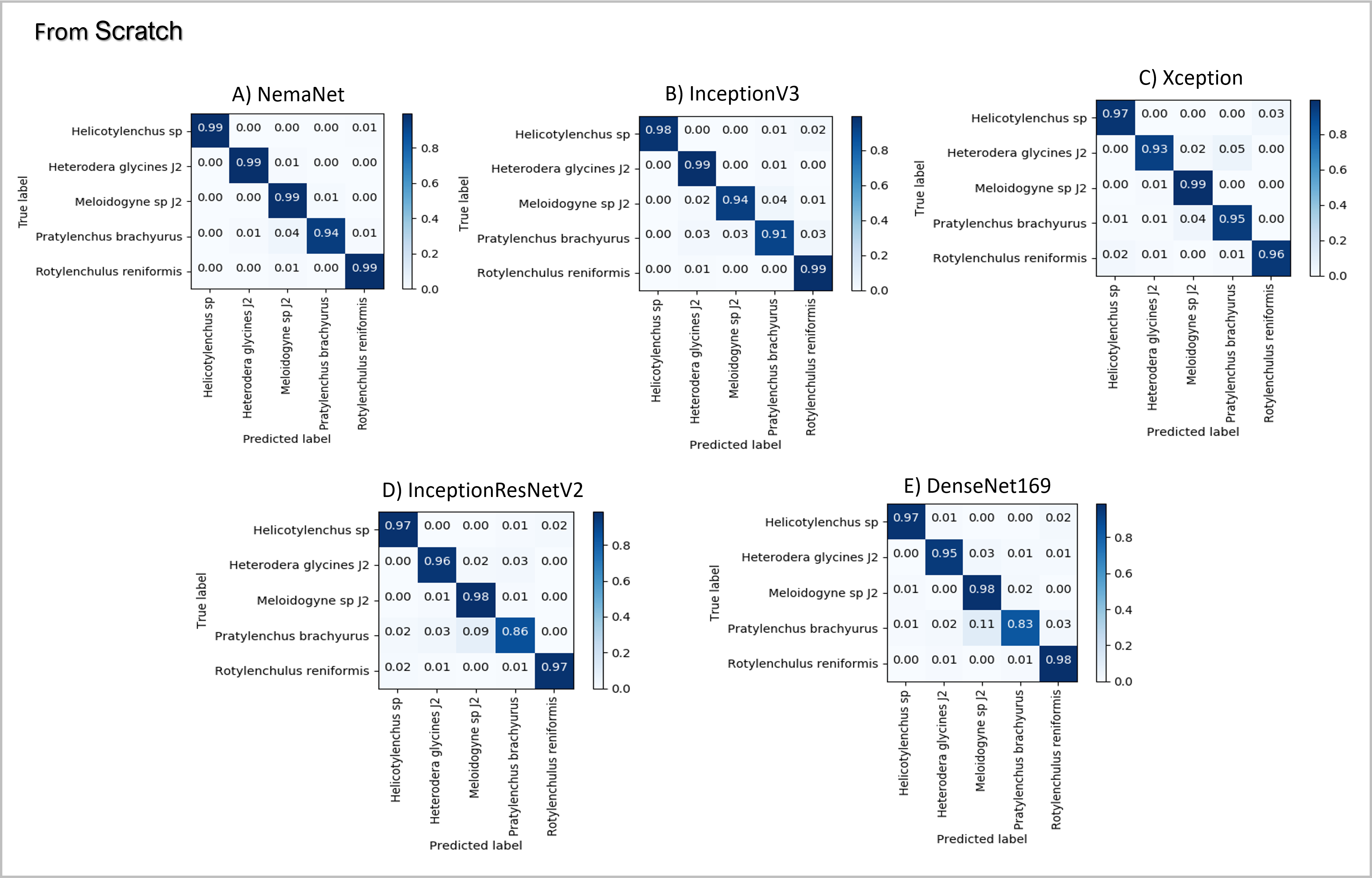}
\caption{Normalized Confusion Matrix - From Scratch: \textbf{(A)} NemaNet; \textbf{(B)} InceptionV3; \textbf{(C)} Xception; \textbf{(D)} InceptinResNetV2 and \textbf{(E)} DenseNet169 }\label{fig:MatrixFromScratch}
\end{figure}

\begin{figure}[!ht]
  \centering
  \includegraphics[width=13.5cm]{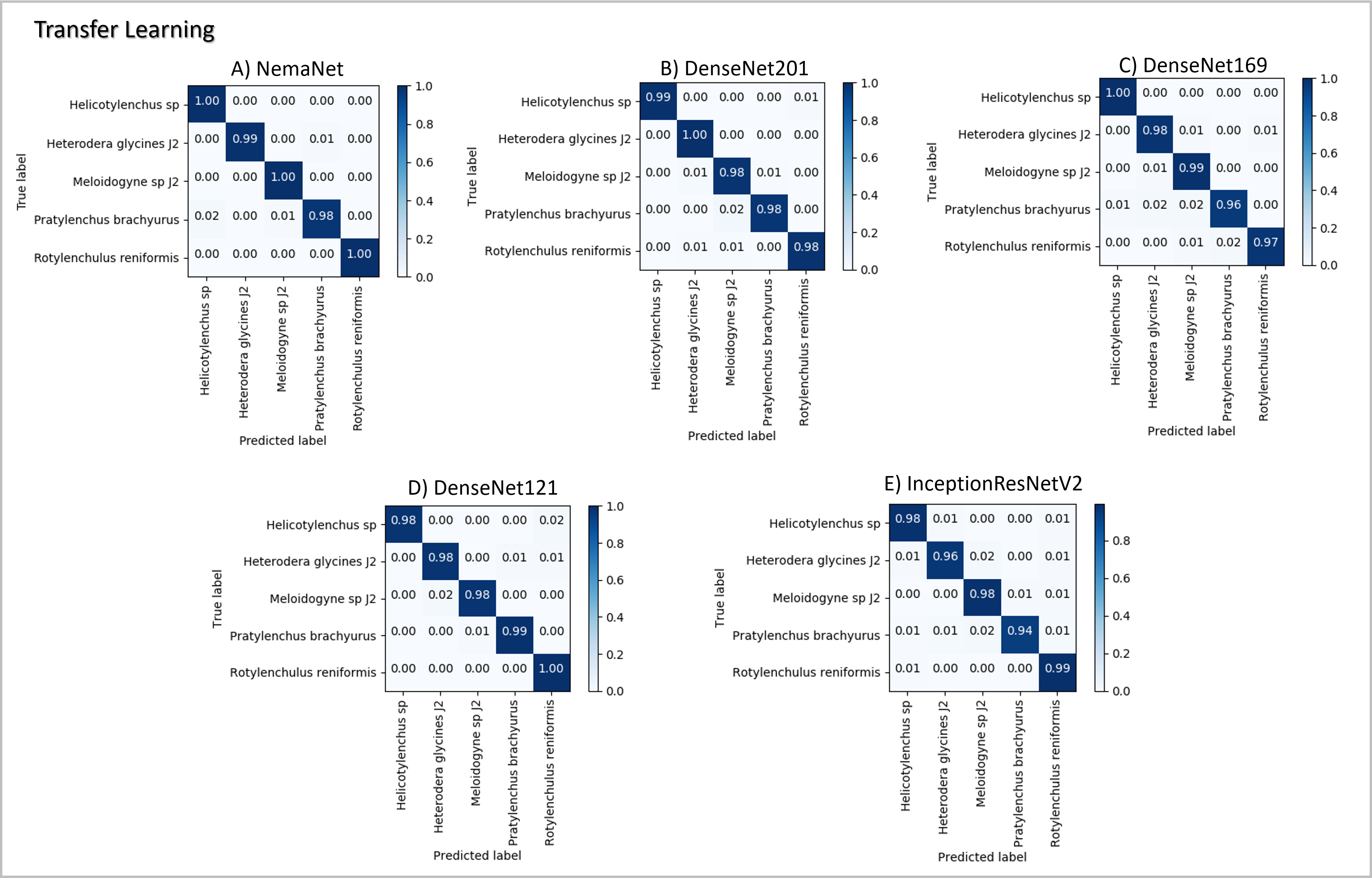}
\caption{Normalized Confusion Matrix - Transfer Learning: \textbf{(A)} NemaNet; \textbf{(B)} DenseNet201; \textbf{(C)} DenseNet169; \textbf{(D)} DenseNet121 and \textbf{(E)} InceptionResNetV2 }\label{fig:MatrixTransfer}
\end{figure}

\begin{figure}[!ht]
  \centering
  \includegraphics[width=12.5cm]{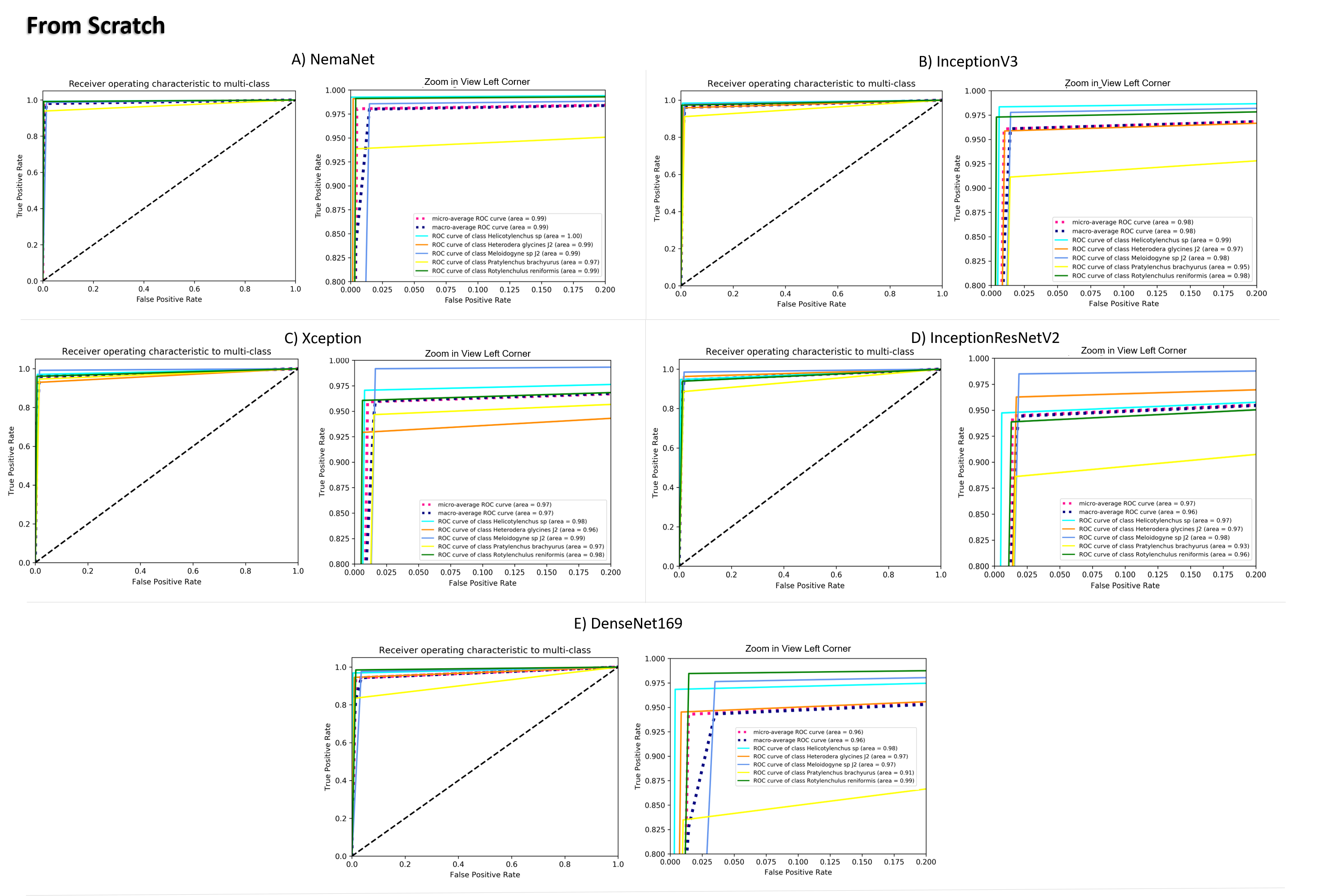}
\caption{Receiver Operating Characteristic (ROC) to MultiClass - From Scratch: \textbf{(A)} NemaNet; \textbf{(B)} DenseNet201; \textbf{(C)} DenseNet169; \textbf{(D)} DenseNet121 and \textbf{(E)} InceptionResNetV2 }\label{fig:ROC_FromScratch}
\end{figure}

\begin{figure}[!ht]
  \centering
  \includegraphics[width=12.5cm]{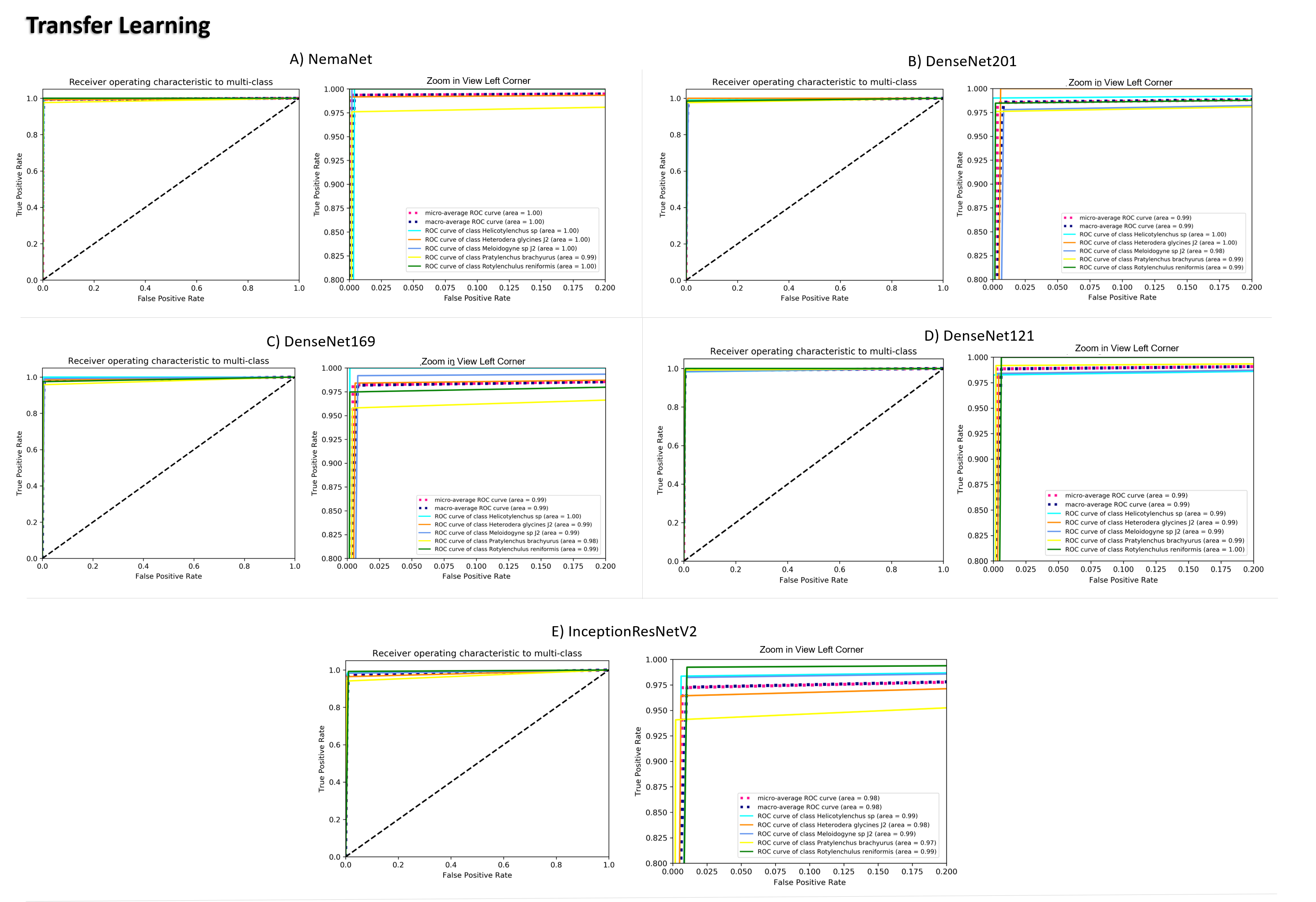}
\caption{Receiver Operating Characteristic (ROC) to MultiClass - Transfer Learning: \textbf{(A)} NemaNet; \textbf{(B)} DenseNet201; \textbf{(C)} DenseNet169; \textbf{(D)} DenseNet121 and \textbf{(E)} InceptionResNetV2 }\label{fig:ROC_Transfer}
\end{figure}

In Figures~\ref{fig:MatrixFromScratch} (FS) and \ref{fig:MatrixTransfer} (TL) we present the confusion matrices normalized for the five best models according to the training strategies used. Additionally, Figures~\ref{fig:ROC_FromScratch} (FS) and \ref{fig:ROC_Transfer} (TL) present the ROC Curve Multiclass calculed for each CNN model.

\section{Discussion}\label{sec:discussion}

Our approach evaluated thirteen popular architectures of convolutional neural networks and proposed a new custom architecture called NemaNet to identify the main nematodes that cause damage to soybeans. The results showed that our proposal obtained the best performance among the other architectures evaluated, both for FS training and for TL.

Among the models of CNNs trained and evaluated using the From Scratch technique, architectures composed of Inception blocks occupy prominent positions among the first five positions in terms of average accuracy, namely InceptionV3, Xception, and InceptionResNetV2. The Xception and InceptionResNetV2 models that use residual connections combined with the Inception architecture performed less than the InceptionV3 network. According to~\cite{ResidualInception:2017} there is a need to customize these architectures to show the gains provided by residual connections. 

The fundamental building block of Inception-style models is the Inception module, of which several different versions exist. The Inception block is equivalent to a subnetwork with four paths. It extracts information in parallel through convolutional layers of varying window shapes and maximum pooling layers~\cite{InceptionV3:2016}. While Inception modules are conceptually similar to convolutions (they are convolutional feature extractors), they empirically appear to be capable of learning richer representations with fewer parameters~\cite{Xception:2017}. 

When we evaluate models trained using TL, architectures composed of dense blocks called DenseNets occupy the first five positions in terms of average accuracy, namely DenseNet201, DenseNet169, and DenseNet121. The DenseNet architecture has several compelling advantages: they alleviate the vanishing-gradient problem, strengthen feature propagation, encourage feature reuse, and substantially reduce the number of parameters~\cite{DenseNet:2017}.

In DenseNet, the classifier uses features of all complexity levels. It tends to give more smooth decision boundaries. It also explains why DenseNet performs well when training data is insufficient. Each layer in DenseNet receives all preceding layers as input, more diversified features, and richer patterns. With training using a pre-trained model in the ImageNet data set, this model leverages features extracted by very early layers directly used by deeper layers throughout the same dense block. This functionality, combined with fine-tuning and the optimization of hyperparameters, increased the predictive capacity of this architecture.

NemaNet explores the different behavior of the two architectures, InceptionV3 and DenseNet121, customizing a model capable of converging with the least number of epochs, with model size and parameter numbers slightly higher than DenseNet121 and presenting greater accuracy. A lot of care was put into ensuring that the model would perform well on high and low-resolution images, enabled by the Inception blocks, and analyze the image representations at different scales. 

 Another critical factor is the occurrence of overfitting in the original DenseNet architectures with FS training. Our customization through the addition of Inception blocks to the DenseNet121 structure guarantees network stability, even in the face of a data set considered as small as the NemaDataSet, used in our experiments.

The stochastic gradient descent optimizer with momentum was used to train all evaluated models. The learning rate is one of the most relevant hyperparameters for a CNN training process and possibly the key to practical and faster training of the network. In our experiments, we chose to use a cyclic learning rate (CLR) method that virtually eliminates the need to adjust the learning rate, achieving almost ideal classification accuracy. Unlike adaptive learning rates, the CLR method essentially does not require any additional calculations\cite{CLR:2017}.

However, it was possible to observe that during the NemaNet training and validation process, there was a sudden variation in the loss values and, consequently, the lower accuracy in the first twenty training epochs. We associate this fact with the hybrid structure of the network, composed of dense blocks and inceptions blocks. This peculiarity makes the CLR method of adjusting the learning rate cause more significant variations since its changes occur cyclically with each batch, instead of a non-cyclical learning rate constant or changes each epoch.
 
Notably, when we use TL in the training and validation process, it is possible to notice that most of the evaluated models' behavior tends to be an appropriate fitting. In this respect, when evaluating NemaNet and its proposed hybrid topology, we use pre-trained weights on ImageNet only for the standard part of the DenseNet121 structure. Bearing in mind that inception blocks are customized and adapted to function as a parallel and auxiliary structure, there is no possibility of reusing knowledge for these blocks, so we consider that NemaNet partially uses the benefits of TL.

Our strategy using a partial TL is successful when we compare the training and validation process. We realize that the convergence capacity of our approach is superior to the other models evaluated.

Another important factor is that all trained models and evaluated using TL performed better than the FS training. Thus, even in the face of the peculiarities of microscopic images and the similar morphologies of the species of phytonematodes, the discriminatory behavior of CNN models proved an efficient approach for identifying these pathogens.

\section{Conclusions and Further Works}\label{sec:conclusion}

In this work, we present a public and opensource dataset called NemaDataset with the five main species of phytonematodes that cause damage to soybean crops. Besides, we trained and evaluated thirteen CNN models using the NemaDataset, representing the state of the art of object classification and recognition in the computer vision research area. All models were compared with our proposed topology NemaNet.

Through the used metrics, our experiments demonstrate an average variation in precision for FS training (79.10\% to 96.99\%) and TL (92.09 \% to 98.88 \%). Our model, called NemaNet, achieved the best results in the two training and validation strategies used.

In the general average calculated for each model, for FS training, NemaNet reached 96.99\% accuracy, while the best evaluation fold reached 98.03\%. As for TL training, the average accuracy reached 98.88\%, while the best evaluation fold reached 99.34\%.

In general, our results are encouraging and can guide future approaches that intend to explore the challenges and gaps in the identification and classification of phytonematodes using a customized architecture. However, it should be noted that we did not evaluate NemaNet in other datasets, nor did we have the opportunity to validate its performance with a data set with a larger number of classes. In the literature, it was not possible to find other phytonematodes datasets that would allow a comparison and performance evaluation.

In this way, in the future, we intend to increase the number of species (classes) of phytonematodes in our dataset and, consequently, increase the number of samples for each category, expanding its representation capacity. Also, descriptive images with taxonomic keys will be added to improve the ability to extract characteristics for species phenotyping.

Other optimizations in our topology model will be implemented, this time using the Neural Architecture Research (NAS) techniques. There are many approaches related to architectural search spaces, optimization algorithms, and methods for evaluating candidate architectures. In this respect, when we define a search space with the Inception and dense blocks, we believe in the possibility of composing even more efficient architectural arrangements.

\section*{Acknowledgment}

The authors also express their gratitude and acknowledge the support of NVIDIA Corporation with the donation of the GPU Titan Xp used for this research and the technical support of the Laboratory of Phytopathology of the Federal University of the State of Mato Grosso - Campus of  Araguaia.

\bibliographystyle{unsrt}  
\bibliography{references}  

\end{document}